\documentclass[%
 reprint,
 amsmath,amssymb,amsfonts,
 aps,
]{revtex4-2}

\usepackage[dvipdfmx]{graphicx}
\usepackage{dcolumn}% Align table columns on decimal point
\usepackage{bm}% bold math
\usepackage{color}
\usepackage{url}
\usepackage{tabularx}
\usepackage{multirow}
\usepackage{mathtools}
\usepackage[dvipsnames]{xcolor}

\def\mat#1{#1}
\def\vect#1{\mbox{\boldmath $#1$}}

\newcommand{\LCE}{LCE}
\newcommand{\NMI}{NMI}
\newcommand{\SBM}{SBM}
\newcommand{\ORGM}{ORGM}
\newcommand{\SupplementaryMaterials}{Supplemental Material}
\newcommand{\papertitle}{Consistency between ordering and clustering methods for graphs}

\newcounter{figcountSI}

\newcommand{\figcountSI}{\refstepcounter{figcountSI}}
\newcounter{tabcountSI}

\newcommand{\tabcountSI}{\refstepcounter{tabcountSI}}
\newcounter{seccountSI}

\newcommand{\seccountSI}{\refstepcounter{seccountSI}}

\begin{document}

% Use the \preprint command to place your local institutional report
% number in the upper righthand corner of the title page in preprint mode.
% Multiple \preprint commands are allowed.
% Use the 'preprintnumbers' class option to override journal defaults
% to display numbers if necessary
%\preprint{}

%Title of paper
%\title{Permutation test for clustering and seriation}
\title{\papertitle}

% repeat the \author .. \affiliation  etc. as needed
% \email, \thanks, \homepage, \altaffiliation all apply to the current
% author. Explanatory text should go in the []'s, actual e-mail
% address or url should go in the {}'s for \email and \homepage.
% Please use the appropriate macro foreach each type of information

% \affiliation command applies to all authors since the last
% \affiliation command. The \affiliation command should follow the
% other information
% \affiliation can be followed by \email, \homepage, \thanks as well.
\author{Tatsuro Kawamoto}
%\email[]{Your e-mail address}
%\homepage[]{Your web page}
%\thanks{}
%\altaffiliation{}
\affiliation{Artificial Intelligence Research Center, \\
  National Institute of Advanced Industrial Science and Technology, 
  Tokyo, Japan }

\author{Masaki Ochi}
%\email{ochi@iis.u-tokyo.ac.jp}
% \affiliation{Department of  Physics, The  University of Tokyo,\\
% 5-1-5 Kashiwanoha, Kashiwa, 277-8574, Chiba, Japan}
 \affiliation{Department of  Physics, The  University of Tokyo, Chiba, Japan}

\author{Teruyoshi Kobayashi}
%\email[]{Your e-mail address}
%\homepage[]{Your web page}
%\thanks{}
%\altaffiliation{}
\affiliation{Department of Economics, Kobe University, Kobe, Japan}

%Collaboration name if desired (requires use of superscriptaddress
%option in \documentclass). \noaffiliation is required (may also be
%used with the \author command).
%\collaboration can be followed by \email, \homepage, \thanks as well.
%\collaboration{}
%\noaffiliation

\date{\today}

\begin{abstract}
A relational dataset is often analyzed by optimally assigning a label to each element through clustering or ordering. 
While similar characterizations of a dataset would be achieved by both clustering and ordering methods, the former has been studied much more actively than the latter, particularly for the data represented as graphs. 
This study fills this gap by investigating methodological relationships between several clustering and ordering methods, focusing on spectral techniques. 
Furthermore, we evaluate the resulting performance of the clustering and ordering methods. 
To this end, we propose a measure called the \emph{label continuity error}, which generically quantifies the degree of consistency between a sequence and partition for a set of elements. 
Based on synthetic and real-world datasets, we evaluate the extents to which an ordering method identifies a module structure and a clustering method identifies a banded structure. 
\end{abstract}

% insert suggested keywords - APS authors don't need to do this
%\keywords{}

%\maketitle must follow title, authors, abstract, and keywords
\maketitle

\section{Introduction \label{Introduction}}
Identifying macroscopic connection patterns in graphs is a major challenge in network science. 
A number of algorithms have been proposed to extract different features, such as community structure \cite{SCHAEFFER2007,Fortunato2010,Fortunato2016}, hierarchical community structure \cite{clauset2008hierarchical,PeixotoPhysRevX2014}, core-periphery structure \cite{Rombach2014}, nested structure \cite{MARIANI2019}, and banded structure \cite{Liiv2010,Behrisch2016,KawamotoKobayashi2021}, to name a few. 

\begin{figure}[t!]
  \centering
  \includegraphics[width= 0.75\columnwidth]{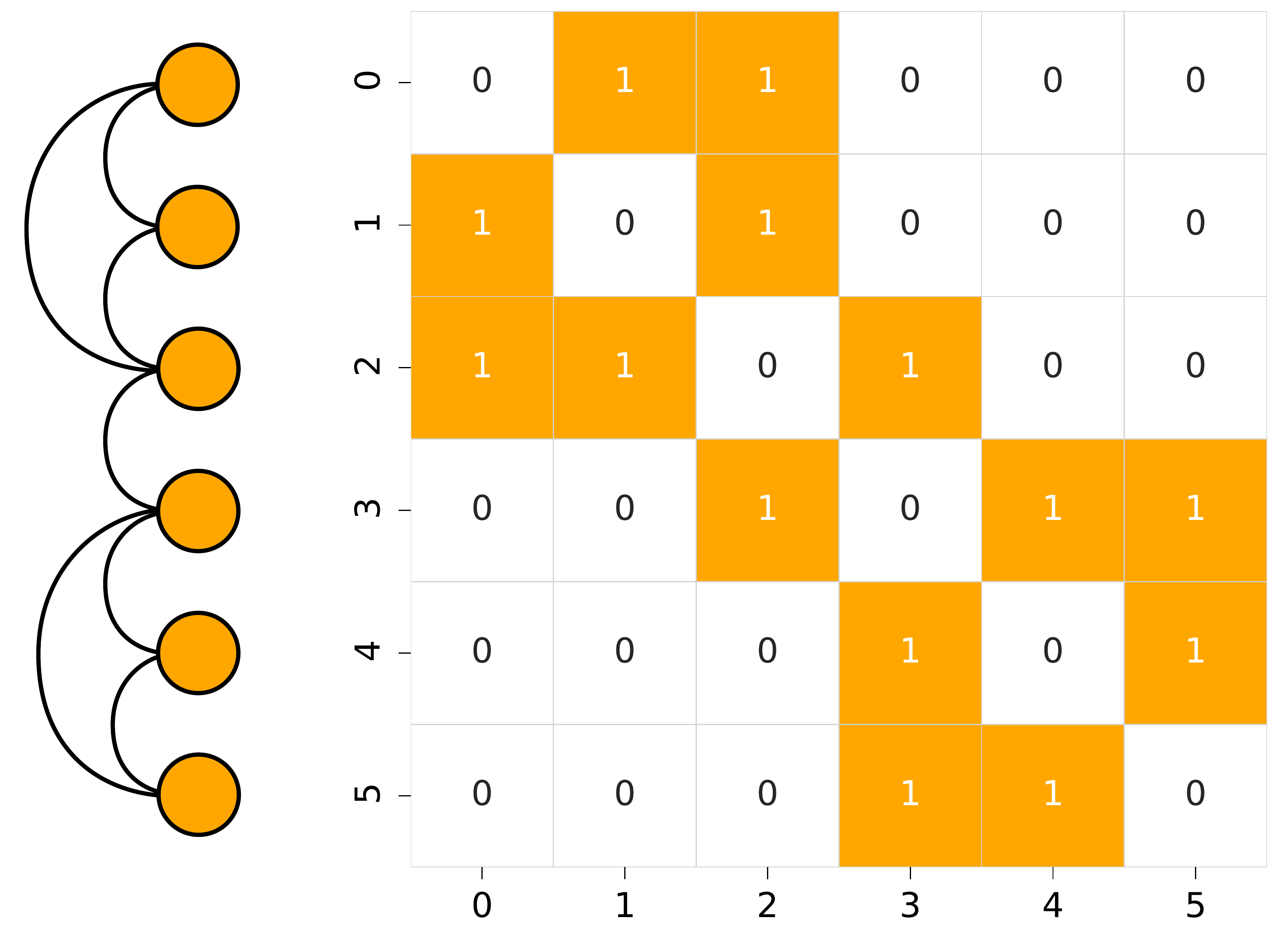}
  \caption{
  Simple example of a graph (left) and its adjacency matrix (right) for identifying a community structure through the optimal ordering of vertices without specifying the group labels. 
  The $(i,j)$ element of the adjacency matrix is one (highlighted) when vertices $i$ and $j$ are connected, and zero (not highlighted) otherwise.
	}
  \label{fig:SimpleExample}
\end{figure}

When a graph consists of subgraphs in each of which vertices are densely connected, the graph structure is referred to as a community structure. 
A common approach for extracting a community structure is the partitioning of graphs, termed community detection \cite{Fortunato2010,Fortunato2016} or graph clustering \cite{SCHAEFFER2007}. 
In this approach, an algorithm assigns a group label to each vertex such that vertices with the same group label are densely connected. 
Alternatively, we may also identify densely connected vertices through an \emph{ordering} method that infers the optimal ordering of vertices such that vertices close to each other in the sequence are densely connected. 
The corresponding optimization problems are collectively termed the minimum linear arrangement \cite{Harper1964,Chung1984,Seitz2010} or envelope reduction \cite{Barnard95,DingHe2004}, and the inferred structural property is called a banded structure or sequentially local structure \cite{KawamotoKobayashi2021}. 
As exemplified in Fig.~\ref{fig:SimpleExample}, the densely connected vertices are clearly detected by appropriately visualizing the graph and the adjacency matrix based on an appropriate vertex ordering. 
 
\begin{figure}[t!]
  \centering
  \includegraphics[width= \columnwidth]{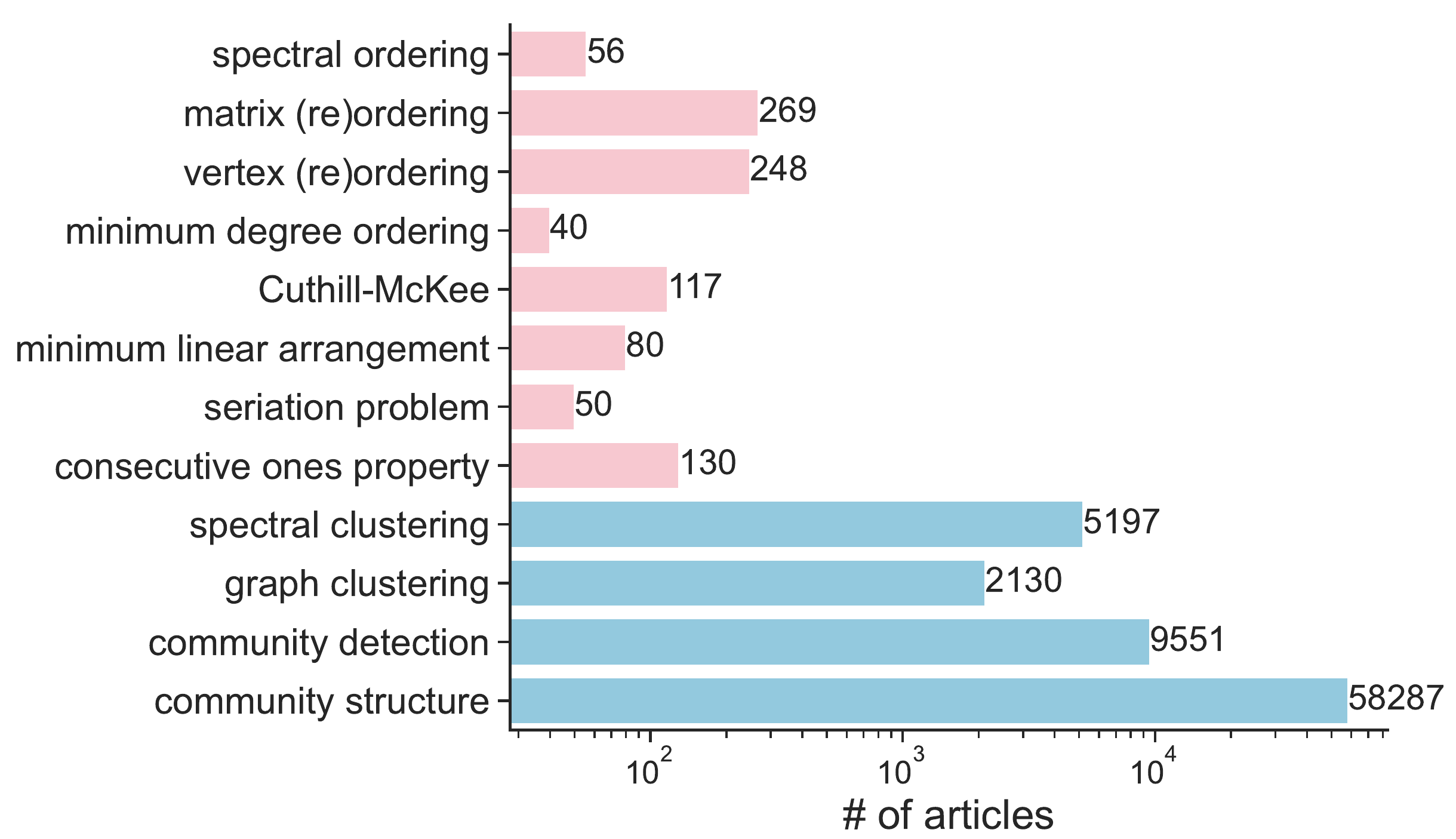}
  \caption{
  Number of articles with a keyword related to the ordering (pink bar) or clustering (blue bar) problems in the title or abstract. 
  The data was collected from Dimensions~\cite{DimensionsURL} on May 30, 2022. 
  %Although the actual number of articles related to each subject should be a lot more, it reflects the abundance of studies in each field.
	}
  \label{fig:publication_num_citations}
\end{figure}

Despite the similarity between these two approaches, the clustering problem has received considerable attention in the literature. 
Figure \ref{fig:publication_num_citations} shows the number of articles with keywords that represent ordering (pink bars) or clustering (blue bars) problems. 
Most of the keywords for ordering problems represent more general matrix ordering problems rather than vertex ordering problems for graphs (i.e., adjacency matrices), whereas the keywords for clustering problems mostly capture problems for graphs. 
Clustering methods have been studied and applied much more actively than ordering methods. 

Spectral methods are popular in both ordering and clustering problems; the former and the latter are respectively termed spectral ordering \cite{Barnard95,DingHe2004,Liiv2010} and spectral clustering \cite{Hagen1992,ShiMalik2000,LuxburgTutorial}. 
In both methods, the leading eigenvector(s) of a Laplacian or its variant is used to identify the optimal ordering or clustering of vertices. 
Specifically, when a graph is partitioned into two groups based on the sorting of the eigenvector elements \cite{Hagen1992}, the result of spectral clustering is generally consistent with the vertex sequence inferred by spectral ordering. 

However, spectral ordering and clustering algorithms are not generally consistent. 
For instance, when graphs are partitioned into more than two groups, it is common to employ the K-means algorithm \cite{kmeans} on $K (>2)$ leading eigenvectors to achieve a $K$-way partitioning \cite{LuxburgTutorial}. 
By contrast, to identify the optimal vertex sequence using the spectral ordering method, we always use the eigenvector associated with the second leading eigenvalue. 
Therefore, it is nontrivial to determine the extent to which the two methods are quantitatively consistent. 
Even when we partition a graph into two groups, the result of spectral clustering may not be consistent with the vertex sequence obtained by spectral ordering when the K-means algorithm is used to obtain a partition. 

We conduct a systematic investigation to evaluate the consistency between the spectral ordering and clustering methods. 
We first introduce a generic measure, referred to as the \textit{label continuity error} (LCE), to quantify the difference between a sequence and partition for a set of elements (e.g., vertices of graphs). 
Intuitively, a sequence and partition are more consistent with each other if, for a given number of groups, the group label flips less often when following the elements in the specified order. 
We provide a more precise definition in the next section. 
Although we use this measure throughout the study, it is not the only method of quantifying consistency; we will revisit this point in Sec. ~\ref{sec:Discussion}. 

There are also several modern spectral clustering algorithms with unexplored ordering counterparts. 
These include the methods based on the modularity matrix~\cite{newman2006finding,Newman2006politicalbooks}, Bethe Hessian \cite{Saade2014,RevisitingBetheHessian2019}, and regularized Laplacian \cite{Chaudhuri2012,Amini2013,QinRohe2013,Joseph2016,ZhangRohe2018}. 
To fill this gap, we show how spectral ordering algorithms can be derived from optimization problems using the matrices on which these modern spectral clustering methods are based. 
Spectral ordering problems based on these matrices are formulated as variants of the classical spectral ordering problem \cite{Barnard95,DingHe2004} with different penalty terms and/or constraints.

The remainder of this paper is organized as follows. 
Section~\ref{sec:ErrorRate} formally introduces the {\LCE} to quantitatively evaluate the consistency between ordering and clustering methods and examine its properties.
Section~\ref{sec:SpectralMethods} formulates spectral ordering methods corresponding to existing spectral clustering methods for graphs. 
Using the {\LCE} introduced in Sec.~\ref{sec:ErrorRate} and the methods formulated in Sec.~\ref{sec:SpectralMethods}, we analyze the consistency between spectral ordering and clustering methods using synthetic and real-world networks in Sec.~\ref{sec:PerformanceAnalysis}. 
Finally, Sec.~\ref{sec:Discussion} discusses the results of this study.

\section{Label continuity error \label{sec:ErrorRate}}
Let $G(V, E)$ be a graph, where $V = \{v_{1}, \dots, v_{N}\}$ is the vertex set and $E$ is the edge set. 
We assume that every vertex in the graph is distinguishable and let $\mathcal{I} = \{1, \dots, N\}$ be the ordered set indicating the original sequence of the vertices which corresponds to the subcripts in $\{v_{1}, \dots, v_{N}\}$. 
For vertex $v_{i} \in V$ ($i \in \mathcal{I}$), we denote $\pi(i) = \pi_{i} \in \{1, \dots, N\}$ as the index after permutation (i.e., we use $\pi$ as both a mapping and a variable) and $\vect{\pi} = \{\pi(i) | i \in \mathcal{I} \}$ as the reordered sequence of the vertices. 
Similarly, we denote $\sigma(i) = \sigma_{i} \in \{1, \dots, K\}$ as the group label of vertex $v_{i}$ and $\vect{\sigma} = \{\sigma(i) | i \in \mathcal{I} \}$ as the partition of the vertex set. 
We also denote $V_{k} = \{v_{i} | \sigma_{i}=k, i \in \mathcal{I}\}$ and $N_{k} = |V_{k}|$ for group $k$ (we let $\{N_{1}, \dots, N_{K}\} =: \{ N_{k} \}$). 
Throughout this study, $\hat{\vect{\pi}}$ and $\hat{\vect{\sigma}}$ represent the inferred sequence and partition using algorithms, respectively.
We denote $d_{i}$ for the degree of vertex $v_{i}$. 
%Let $G(V, E)$ be a graph, where $V$ is the vertex set and $E$ is the edge set. 
%We assume that every vertex in the graph is distinguishable and let $\mathcal{I} = \{1, \dots, N\}$ be the original sequence of the vertex set $V$. 
%For vertex $i \in \mathcal{I}$, we denote $\pi(i) = \pi_{i} \in \{1, \dots, N\}$ as the index after permutation (i.e., we use $\pi$ as both a mapping and a variable) and $\vect{\pi} = \{\pi(i) | i \in \mathcal{I} \}$ as the reordered sequence of the vertices. 
%Similarly, we denote $\sigma(i) = \sigma_{i} \in \{1, \dots, K\}$ as the group label of vertex $i$ and $\vect{\sigma} = \{\sigma(i) | i \in \mathcal{I} \}$ as the partition of the vertex set. 
%We also denote $V_{k} = \{i | \sigma_{i}=k, i \in \mathcal{I}\}$ and $N_{k} = |V_{k}|$ for group $k$ (we let $\{N_{1}, \dots, N_{K}\} =: \{ N_{k} \}$). 
%Throughout this study, $\hat{\vect{\pi}}$ and $\hat{\vect{\sigma}}$ represent the inferred sequence and partition using algorithms, respectively.

\subsection{Definition}
\begin{figure}[t!]
  \centering
  \includegraphics[width= 0.95\columnwidth]{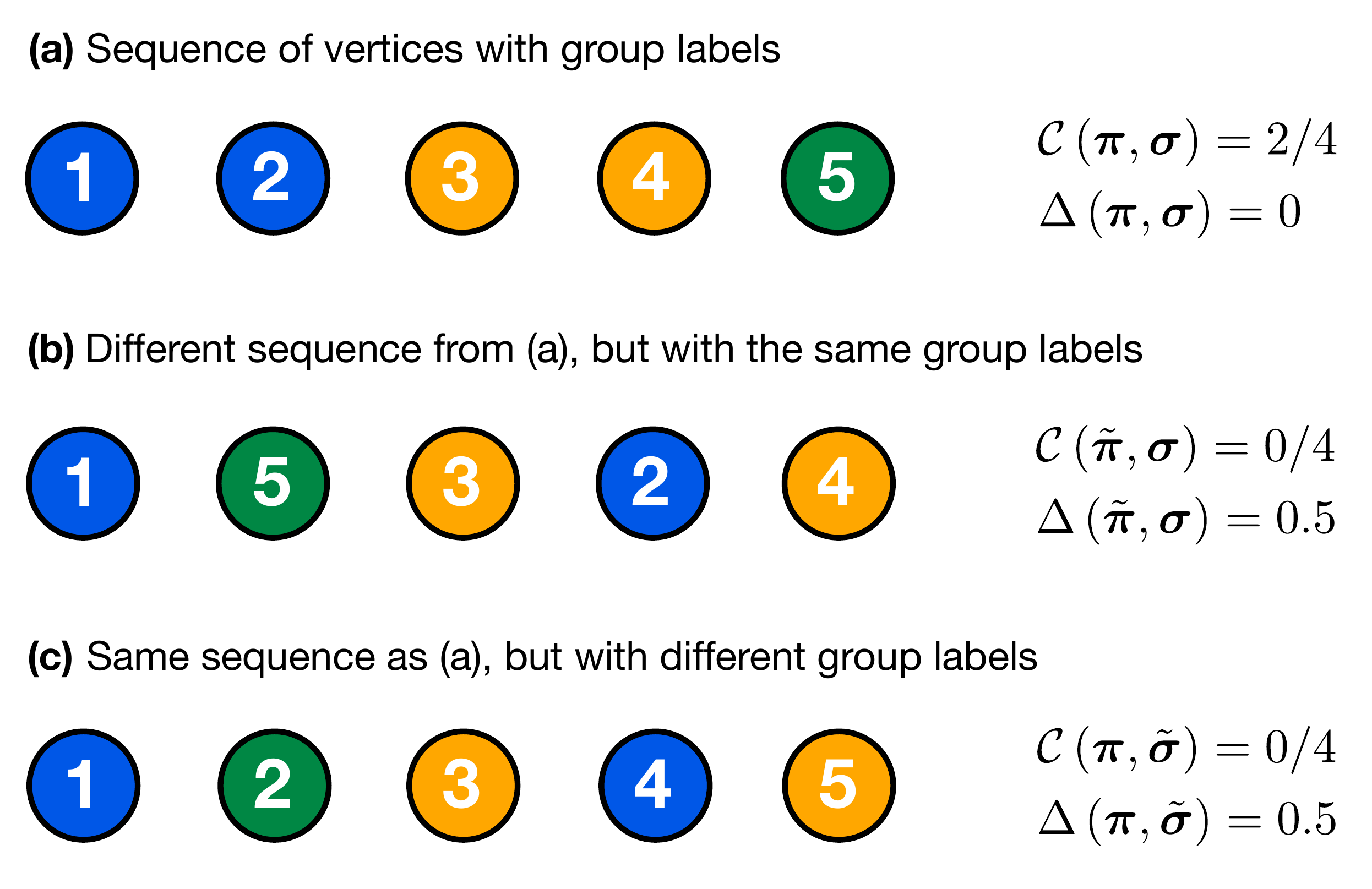}
  \caption{
  Examples of the label continuity $\mathcal{C}\left( \vect{\pi}, \vect{\sigma} \right)$ and the label continuity error $\Delta\left( \vect{\pi}, \vect{\sigma} \right)$ for different sequences and partitions for the same vertex set. 
  The number on each vertex represents the original index of the vertex. 
	}
  \label{fig:SchematicLCE}
\end{figure}

%We introduce a measure to indicate the consistency between a sequence $\vect{\pi}$ and partition $\vect{\sigma}$. 
%We define that the sequence $\vect{\pi}$ is consistent with $\vect{\sigma}$ if the group labels are consecutive along the sequence. 
We introduce a measure to quantify the consistency between a sequence $\vect{\pi}$ and partition $\vect{\sigma}$.
We define the sequence $\vect{\pi}$ as consistent with $\vect{\sigma}$ if vertices with the same group label are maximally adjacent to each other in the sequence $\vect{\pi}$. 
For instance, if the original indices $\mathcal{I}$ are consistent with group labels $\vect{\sigma}$, 
\begin{align}
\sum_{i=1}^{N-1} \delta\left( \sigma(i), \sigma(i+1) \right)
\label{RawOverlap}
\end{align}
is maximized, where $\delta(a,b)$ represents the Kronecker delta; Fig.~\ref{fig:SchematicLCE}(a) presents an example. 
To evaluate the consistency between $\vect{\pi}$ and $\vect{\sigma}$, we introduce a measure that we refer to as the \textit{label continuity}, defined by 
\begin{align}
\mathcal{C}\left( \vect{\pi}, \vect{\sigma} \right) = 
\frac{1}{N-1}\sum_{i^{\prime}=1}^{N-1} \delta\biggl( \sigma\left( \pi^{-1}(i^{\prime}) \right), \sigma\left( \pi^{-1}(i^{\prime}+1) \right) \biggr),
\label{ContinuityDef}
\end{align}
where $\pi^{-1}$ is the inverse mapping of $\pi$ and $i^{\prime}$ is the index label after the permutation; that is, $\pi^{-1}(i^{\prime})$ is the label $i$ in the original indices satisfying $\pi(i) = i^{\prime}$. 
The number of times that the group labels are flipped when following the vertices in the order of $\vect{\pi}$ is expressed as $(N-1)(1-\mathcal{C}\left( \vect{\pi}, \vect{\sigma} \right))$, for which the group labels must be flipped at least $K-1$ times. 
Considering this feature, we define the \textit{label continuity error} ({\LCE}) as 
\begin{align}
\Delta\left( \vect{\pi}, \vect{\sigma} \right) = 
1 - \frac{K-1}{N-1} - \mathcal{C}\left( \vect{\pi}, \vect{\sigma} \right).
\label{LabelContinuityError}
\end{align}
Hereafter, we abbreviate $\mathcal{C}\left( \vect{\pi}, \vect{\sigma} \right)$ and $\Delta\left( \vect{\pi}, \vect{\sigma} \right)$  as $\mathcal{C}$ and $\Delta$, respectively, as long as there is no possibility of confusion. 

For a given partition $\vect{\sigma}$ and different vertex sequences, we can evaluate which vertex sequence is more consistent with $\vect{\sigma}$ using the {\LCE} (e.g., Figs.~\ref{fig:SchematicLCE}(a) and \ref{fig:SchematicLCE}(b)). 
Similarly, for a given sequence $\vect{\pi}$ and different partitions, we can also evaluate which partition is more consistent with $\vect{\pi}$, keeping the group sizes $\{ N_{k} \}$ fixed (e.g., Figs.~\ref{fig:SchematicLCE}(a) and \ref{fig:SchematicLCE}(c)).
%For a given partition $\vect{\sigma}$, we can compare the extents to which different vertex sequences are consistent with $\vect{\sigma}$ using the {\LCE}. 
%Similarly, for a given sequence $\vect{\pi}$, we can also compare the extents to which different partitions are consistent with $\vect{\pi}$, keeping the number of groups fixed.

\subsection{Properties of the {\LCE}}
The {\LCE} can take only small values when the number of groups $K$ is very small or large. 
For example, it is obvious that $\Delta$ is zero when $K=1$ or $K=N$. 
In other words, the resolution of the {\LCE} is low in such regions. 
Moreover, this property would depend on the distribution of the group sizes $\{ N_{k} \}$. 
In this section, we quantify these intuitions. 

The minimum value of $\Delta$ is zero by construction. 
The maximum value of $\Delta$ is obtained when labels are flipped the maximum number of times.
The maximization of $\Delta$ by optimizing the sequence $\vect{\pi}$, given an arbitrary partition $\vect{\sigma}$ with $\{ N_{k} \}$, is equivalent to maximizing $\Delta$ by optimizing partition $\vect{\sigma}$ (constrained to $\{ N_{k} \}$) for a given sequence $\vect{\pi}$. We denote the maximum by $\max \Delta$ as 
\begin{align}
\max_{\vect{\pi}} \Delta\left( \vect{\pi}, \vect{\sigma} \right) 
= \max_{\vect{\sigma} (\{ N_{k} \})} \Delta\left( \vect{\pi}, \vect{\sigma} \right) 
= \max \Delta. 
\end{align}
As derived in Appendix \ref{sec:MaximumLCE}, we have 
\begin{align}
\max \Delta 
= \begin{cases}
\frac{2\left( N - \max_{k} N_{k} \right)}{N-1} - \frac{K-1}{N-1} & \left( \max_{k} N_{k} > \left\lceil \frac{N}{2} \right\rceil \right) \\
1 - \frac{K-1}{N-1} & (\text{otherwise})
\end{cases}, \label{MaxLCE}
\end{align}
where $\lceil \cdot \rceil$ denotes the ceiling function.

We next investigate statistical properties of the {\LCE}. 
First, we calculate the probability $\mathcal{P}(m)$ that the number of times that two consecutive vertices in a sequence have the same group label is $m$, where $m = (N-1) \mathcal{C}\left( \vect{\pi}, \vect{\sigma} \right)$. 
When any sequence realizes at random, we have 
\begin{align}
\mathcal{P}(m) 
= \frac{1}{N!} \sum_{\vect{\pi}^{\prime}} \delta\left( m, (N-1)\mathcal{C}\left( \vect{\pi}^{\prime}, \vect{\sigma} \right) \right), 
\label{RandomSERMicroProb}
\end{align}
where $\vect{\sigma}$ is an arbitrary partition with group sizes $\{ N_{k} \}$ and the sum is over all possible sequences ($|\{ \vect{\pi}^{\prime} \}| = N!$). 
Note that Eq.~(\ref{RandomSERMicroProb}) is also a distribution in which each distinct partition realizes at random. 
This equivalence might sound peculiar because there are only $N!/\prod_{k=1}^{K} N_{k}!$ distinct partitions, whereas there are $N!$ possible sequences.
However, because every distinct partition is overcounted exactly $\prod_{k=1}^{K} N_{k}!$ times in the summation of Eq.~(\ref{RandomSERMicroProb}), the distribution $\mathcal{P}(m)$ is identical for both random sequences and random partitions. 

Although Eq.~(\ref{RandomSERMicroProb}) is a straightforward expression, a strict constraint on $\{ N_{k} \}$ makes analytical calculations complicated. 
Therefore, we instead calculate the distribution of bootstrapped group labels $\vect{\sigma}^{\ast}$ as an approximation. 
That is, we generate a random group assignment $\vect{\sigma}^{\ast}$ by sampling independently from the empirical distribution $\mathrm{Prob}[k] = N_{k}/N$ ($k \in \{1, \dots, K\}$); in other words, we randomly resample group labels from $\vect{\sigma}$ with replacement. 
%Here, we directly evaluate the random partitions instead of random sequences. 
The distribution of group labels $\vect{\sigma}^{\ast}$ is 
\begin{align}
P(\vect{\sigma}^{\ast}) = \prod_{i=1}^{N} \frac{N_{\sigma^{\ast}(i)}}{N}. 
\end{align}
This approximation for random group labels is expected to be accurate if each element in $\{ N_{k} \}$ is sufficiently large. 

Using the bootstrapped group labels, the mean value of $\mathcal{C}$ is obtained as 
\begin{align}
\mathbb{E}\left[ \mathcal{C} \right] 
&= \sum_{\vect{\sigma}^{\ast}} P(\vect{\sigma}^{\ast}) 
\frac{\sum_{i=1}^{N-1} \delta\left( \sigma^{\ast}(i), \sigma^{\ast}(i+1) \right)}{N-1} \notag\\
&= \frac{1}{N-1} \sum_{i=1}^{N-1} \Biggl[ 
\sum_{\sigma^{\ast}_{i}, \sigma^{\ast}_{i+1}} P(\sigma^{\ast}_{i}) P(\sigma^{\ast}_{i+1}) \delta\left( \sigma^{\ast}(i), \sigma^{\ast}(i+1) \right) \notag\\
&\hspace{20pt}\times \prod_{j (\ne i, i+1)} \left( \sum_{\sigma^{\ast}_{j}} P(\sigma^{\ast}_{j}) \right)
\Biggr] \notag\\
&= \sum_{k=1}^{K} \left( \frac{N_{k}}{N} \right)^{2}.
\end{align}
Therefore, the mean value of {\LCE} under random partitioning is
\begin{align}
\overline{\Delta}\left( \{ N_{k} \} \right) := 
\mathbb{E}\left[ \Delta \right] = 
\frac{N-K}{N-1} - \sum_{k=1}^{K} \left( \frac{N_{k}}{N} \right)^{2}.
\label{MeanLabelContinuityError}
\end{align}
As the {\LCE} does not practically become greater than $\overline{\Delta}\left( \{ N_{k} \} \right)$, this mean value is a more meaningful reference value than Eq.~(\ref{MaxLCE}) as the upper bound. 

%When the number of vertices is sufficiently large and the number of groups is $O(1)$, $N/K \gg 1$, the value of $\Delta$ concentrates to $\mathbb{E}\left[ \Delta \right]$ according to the law of large numbers. 
%However, when $N/K = O(1)$, the fluctuation of $\Delta$ may not be negligible. 
We can also derive the variance $\mathrm{Var}[\Delta]$ (the derivation is shown in Appendix \ref{sec:VarSER}) as 
\begin{align}
\mathrm{Var}[\Delta] 
&= \frac{1}{N-1} \sum_{k=1}^{K} \left( \frac{N_{k}}{N} \right)^{2} 
+ \frac{2(N-2)}{(N-1)^{2}} \sum_{k=1}^{K} \left( \frac{N_{k}}{N} \right)^{3} \notag\\
&\hspace{10pt} -\frac{3N - 5}{(N-1)^{2}} \left( \sum_{k=1}^{K} \left( \frac{N_{k}}{N} \right)^{2} \right)^{2}, 
\end{align}
showing that $\Delta$ converges to $\mathbb{E}\left[ \Delta \right]$ by the law of large numbers. 
Furthermore, in Appendix \ref{sec:NullDistribution}, we show that the probability distribution is asymptotically normal when the group sizes are equal and $K=O(1)$, implying that higher-order moments will vanish. 

\begin{figure}[t!]
  \centering
  \includegraphics[width= \columnwidth]{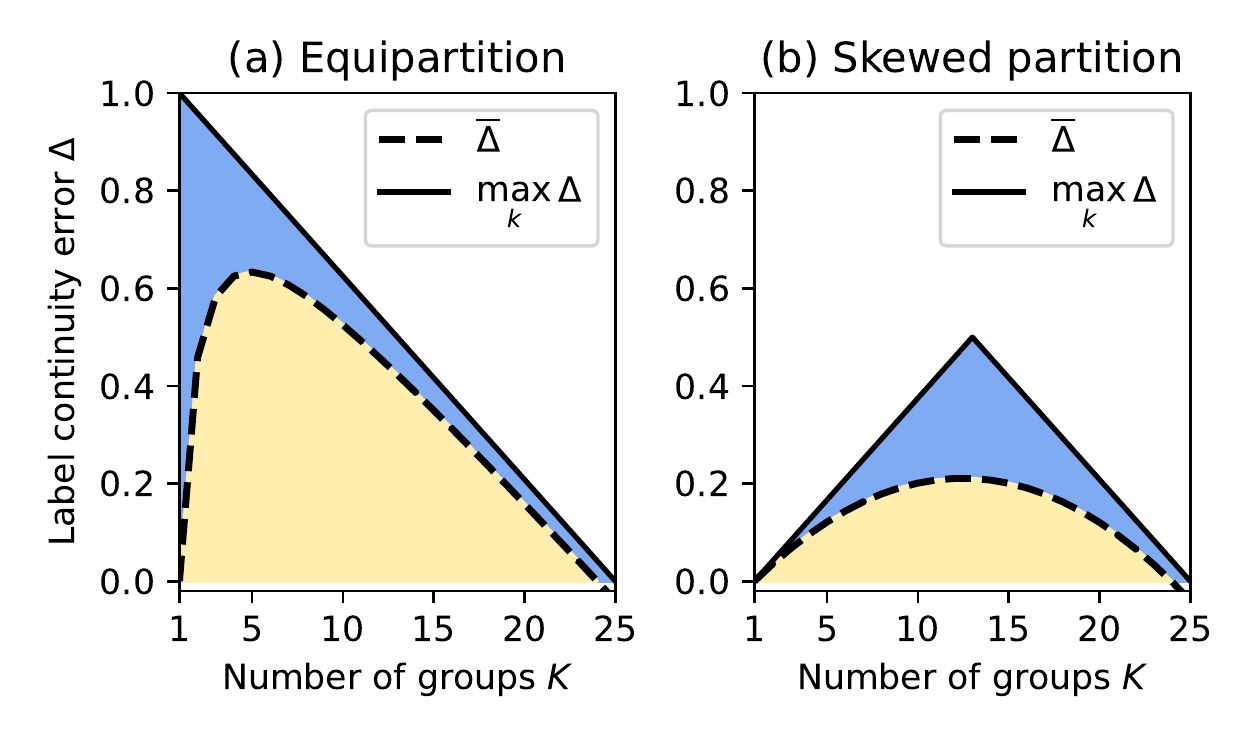}
  \caption{
Upper bound $\max \Delta$ (solid line) of the {\LCE} and bootstrap estimate of the mean values $\overline{\Delta}$ (dashed line) of the {\LCE} under a random sequence as functions of the number of groups $K$: (a) an equipartition (i.e., $N_{k}/N = 1/K$ for any $k \in \{1, \dots, K\}$) and (b) a skewed partition (i.e., $N_{k}=1$ $(2 \le k \le K-1)$ and $N_{1} = N-K+1$). 
  We set $N=25$ in these examples. 
Continuous approximations of $\max \Delta$ and $\overline{\Delta}$ are shown to highlight their dependency on $K$. 
	}
  \label{fig:convexity}
\end{figure}

Let us summarize the results we obtained in this section. 
As the number of groups $K\, (>1)$ increases, the upper bound of the {\LCE} ($\max \Delta$ in Eq.~(\ref{MaxLCE})) decreases monotonically as long as the partitions are not highly skewed, i.e., $\max_{k} N_{k} < \left\lceil N/2 \right\rceil$. 
However, as illustrated in Fig.~\ref{fig:convexity}, the {\LCE} for a random sequence ($\overline{\Delta}$ in Eq.~(\ref{MeanLabelContinuityError})) is a convex function with respect to $K$. 
When $K$ is small, the {\LCE} increases because the chance for label flips increases, while the {\LCE} decreases owing to the increase in the minimum number of label flips. 
For equipartitioning (Fig.~\ref{fig:convexity}(a)), the mean {\LCE} $\overline{\Delta}$ is peaked at an integer of approxmately $K=\sqrt{N-1}$. 
As a partition becomes more skewed (Fig.~\ref{fig:convexity}(b)), $\max \Delta$ and $\overline{\Delta}$ are peaked at larger values of $K$. 
Therefore, when evaluating the {\LCE}, we must implement appropriate normalizations. 

In this study, we focus on comparing partitions with the same number of groups $K$. 
In Appendix~\ref{sec:LCENested}, however, we discuss nested partitions (subpartitions of another partition) as an example in which different partitions have different numbers of groups.

\section{Spectral ordering methods \label{sec:SpectralMethods}}
In this section, we describe variants of spectral ordering methods using different matrices. 
After reviewing the derivation of the standard methods based on the unnormalized and normalized Laplacians, we show how spectral ordering problems can be formulated with the modularity matrix, regularized Laplacian, and Bethe Hessian.

\subsection{Unnormalized Laplacian \label{sec:UnnormalizedLaplacian}}
Spectral ordering is derived as a continuous relaxation of the discrete optimization problem called envelope reduction \cite{Barnard95}. 
This problem optimizes the vertex sequence $\vect{\pi}$ such that each connected pair of vertices is located close to each other in the sequence. 
To this end, the following objective function is considered: 
\begin{align}
H_{2}\left( \vect{\pi}; \mat{A} \right) = \frac{1}{2} \sum_{i,j} A_{ij} \left( \pi_{i} - \pi_{j} \right)^{2}, \label{ObjectiveFunction_H2}
\end{align}
which is the sum of squared distances $\left( \pi_{i} - \pi_{j} \right)^{2}$ with respect to the set of connected vertices. 
The sequence that minimizes this function is the solution to envelope reduction. 

As the minimization of Eq.~(\ref{ObjectiveFunction_H2}) is not computationally feasible, we consider its continuous relaxation. 
That is, we represent $\vect{\pi}$ using a continuous vector $\vect{x} \in \mathbb{R}^{N}$. 
However, if we simply replace $\vect{\pi}$ with $\vect{x}$, $\vect{x} = \vect{0}$ would be the trivial minimizer of $H_{2}\left( \vect{x}; \mat{A} \right)$.
Thus, we constrain $\vect{x}$ such that $\sum_{i=1}^{N} x^{2}_{i}$ is a positive constant (i.e., the spherical constraint) to reflect the fact that $\sum_{i=1}^{N} \pi^{2}_{i}$ is positive regardless of the choice of sequence. 
Therefore, we consider the minimization of the following function: 
\begin{align}
\frac{1}{2} \sum_{i,j} A_{ij} (x_{i} - x_{j})^{2} 
-\lambda \left( \sum_{i=1}^{N} x_{i}^{2} - 1 \right), 
\label{LagrangeUnnormalizedH2}
\end{align}
where $\lambda$ is the Lagrange multiplier. 
The extremum condition in Eq.~(\ref{LagrangeUnnormalizedH2}) yields the following eigenvalue equation with an eigenvector $\vect{\nu}$: 
\begin{align}
\mat{L} \, \vect{\nu} = \lambda \vect{\nu}. \label{UnnormalizedLaplacianEigenvalueEquation}
\end{align}
Here, $\mat{L} \equiv \mat{D} - \mat{A}$ is the unnormalized (or combinatorial) Laplacian, where $\mat{D} = \mathrm{diag}(d_{1}, \dots, d_{N})$ is the degree matrix ($d_{i} = \sum_{j=1}^{N} A_{ij}$). 
Although we would have a vector proportional to $\vect{1}$ (a vector of ones) as the minimizer of Eq.~(\ref{LagrangeUnnormalizedH2}), which is also the eigenvector associated with the smallest eigenvalue of $\mat{L}$, we cannot infer the optimal sequence from $\vect{1}$ because all the elements are identical. 
Therefore, we exclude vectors proportional to $\vect{1}$, which is equivalent to imposing a perpendicular constraint to $\vect{1}$ in Eq.~(\ref{LagrangeUnnormalizedH2}), i.e., $\sum_{i=1}^{N} x_{i} = 0$. 
Then, the minimizer of the objective function is the eigenvector $\vect{\nu}_{2}$ of $\mat{L}$ associated with the second-smallest eigenvalue. 

The estimate of the optimal sequence $\hat{\vect{\pi}}$ using the spectral ordering method is 
\begin{align}
\hat{\vect{\pi}} = \left\{ \mathrm{rank}(\nu_{2i}) | i \in \mathcal{I} \right\}, \label{DiscretizationH2UnnormalizedLaplacian}
\end{align}
where $\nu_{2i}$ is the $i$th element of $\vect{\nu}_{2}$, and $\mathrm{rank}(\nu_{2i})$ is the index of $\nu_{2i}$ in an array in which the vector elements of $\vect{\nu}_{2}$ are sorted in the ascending or descending order.

\subsection{Normalized Laplacian \label{sec:NormalizedLaplacian}}
A spectral ordering method with the normalized Laplacian was derived in \cite{DingHe2004}. 
Note that the objective function (\ref{ObjectiveFunction_H2}) does not have a periodic boundary condition. 
Therefore, while the distance from one vertex at the end of the sequence to another vertex ranges from $1$ to $N-1$, the distance from the vertex at the middle of the sequence ranges from $1$ to $\lfloor N/2 \rfloor$, where $\lfloor \cdot \rfloor$ is the floor function. 
This implies that when a graph has a vertex with a considerably large degree (i.e., a hub), it is typically more beneficial for the minimization objective to assign such a vertex near the middle of the sequence. 
To incorporate this feature, we replace the spherical constraint in Eq.~(\ref{LagrangeUnnormalizedH2}) with the following ellipsoidal constraint: 
\begin{align}
\sum_{i=1}^{N} d_{i} x^{2}_{i} = \mathrm{const.}, \label{EllipsoidalConstraint}
\end{align}
which tends to restrict $x_{i}$ with a large $d_{i}$ to be relatively small (recall that a variable with a large coefficient typically has relatively small values on an ellipsoid). 
Therefore, Eq.~(\ref{EllipsoidalConstraint}) constrains $\vect{x}$ such that $x_{i}$ of a hub vertex $v_{i}$ is near the origin, and when $\vect{x}$ is discretized, the hub vertices are likely to be located near the middle of the sequence. 
Note also that the mean of $\{x_i\}$ is located at the origin because of the perpendicular constraint $\sum_{i=1}^{N} x_{i} = 0$. 

Consequently, Eq.~(\ref{UnnormalizedLaplacianEigenvalueEquation}) is replaced with the following generalized eigenvalue equation with respect to its second-smallest eigenvalue $\lambda_{2}$:
\begin{align}
\mat{L} \, \vect{\nu}_{2} = \lambda_{2} \mat{D} \vect{\nu}_{2}. \label{NormalizedLaplacianEigenvalueEquation1}
\end{align}
This is equivalent to 
\begin{align}
\mat{\mathcal{L}} \vect{z}_{2} = \lambda_{2} \vect{z}_{2}, \label{NormalizedLaplacianEigenvalueEquation2}
\end{align}
where $\mat{\mathcal{L}} \equiv \mat{D}^{-\frac{1}{2}} \mat{L} \mat{D}^{-\frac{1}{2}}$ is the normalized Laplacian and $\vect{z}_{2} \equiv \mat{D}^{\frac{1}{2}} \vect{\nu}_{2}$. 
As $\vect{\nu}_{2}$ is a continuous relaxation of the sequence $\vect{\pi}$, we estimate the optimal sequence $\hat{\vect{\pi}}$ as 
\begin{align}
\hat{\vect{\pi}} = \left\{ \mathrm{rank}(d^{-1/2}_{i} z_{2i}) | i \in \mathcal{I} \right\}. \label{DiscretizationH2NormalizedLaplacian}
\end{align}

\subsection{Modularity matrix \label{sec:ModularityMatrix}}
The modularity matrix $\mat{Q}$ appears in the spectral clustering method for modularity maximization in community detection \cite{newman2006finding}. 
The matrix element is commonly defined as 
\begin{align}
Q_{ij} = A_{ij} - \frac{d_{i}d_{j}}{2M}, \label{ModularityMatrix}
\end{align}
where $M$ is the total number of edges in the graph. 

To formulate the spectral ordering problem with the modularity matrix, we again consider the objective function $H_{2}\left( \vect{\pi}; \mat{A} \right)$ in the envelope reduction problem and its continuous relaxation with the spherical constraint $\sum_{i=1}^{N} x^{2}_{i} = 1$. 
Herein, we add the following penalty terms to the objective function: 
\begin{align}
\frac{\left( \sum_{i} d_{i} x_{i} \right)^{2}}{2M} - \sum_{i} d_{i} x^{2}_{i}. \label{ModularityPenalization}
\end{align}
The penalty terms ensure that $\{x_i\}$ are ``balanced'' around the origin. 
The first term prohibits $\{x_i\}$ for hub vertices from being located only on the positive or negative side of the real interval $[-1,1]$. 
Owing to the second term, $\{x_i\}$ associated with hub vertices also tend to be away from the origin. 
Therefore, the penalty term Eq.~(\ref{ModularityPenalization}) decreases when $\{x_i\}$ are more symmetrically distributed around the origin. 

Using Lagrange multipliers, the objective function to be minimized is then 
\begin{align}
& \frac{1}{2} \sum_{i,j} A_{ij} (x_{i} - x_{j})^{2} 
+ \frac{\left( \sum_{i} d_{i} x_{i} \right)^{2}}{2M} - \sum_{i} d_{i} x^{2}_{i} 
+\lambda \left( \sum_{i=1}^{N} x_{i}^{2} - 1 \right) \notag\\
%&= -\sum_{i,j} x_{i} \left( A_{ij} - \frac{d_{i}d_{j}}{2M} \right) x_{j} 
&= -\sum_{i,j} x_{i} Q_{ij} x_{j}
+\lambda \left( \sum_{i=1}^{N} x_{i}^{2} - 1 \right).
\label{LagrangeModularityH2}
\end{align}
Here, we do not impose the perpendicular constraint in Eq.~(\ref{LagrangeModularityH2}), because a vector proportional to $\vect{1}$ is not a trivial minimizer. 
The extremum conditions in Eq.~(\ref{LagrangeModularityH2}) yield 
\begin{align}
\mat{Q} \vect{\nu}_{1} = \lambda_{1} \vect{\nu}_{1}, 
\end{align}
where $\lambda_{1}$ represents the largest eigenvalue of $\mat{Q}$ and $\vect{\nu}_{1}$ is the associated eigenvector. 
$\vect{\nu}_{1}$ is the minimizer in Eq.~(\ref{LagrangeModularityH2}) provided that it is not a vector proportional to $\vect{1}$. 
Analogously to Eq.~(\ref{DiscretizationH2UnnormalizedLaplacian}), we estimate the optimal sequence $\hat{\vect{\pi}}$ as 
\begin{align}
\hat{\vect{\pi}} = \left\{ \mathrm{rank}(\nu_{1i}) | i \in \mathcal{I} \right\}. \label{DiscretizationH2ModularityMatrix}
\end{align}

The ellipsoidal constraint enforces $\{x_i\}$ for hub vertices to be concentrated around the origin, whereas the penalty terms (\ref{ModularityPenalization}) enforce them to be evenly distributed at both the positive and negative ends of the real line. 
Therefore, the results of the spectral ordering methods using the normalized Laplacian and modularity matrix are expected to be quite distinct for graphs with heterogeneous degree distributions.

\subsection{Bethe Hessian \label{sec:BetheHessian}}
Bethe Hessian is also a matrix that is originally formulated to perform spectral clustering \cite{Saade2014,RevisitingBetheHessian2019}. 
This method is inspired by the statistical inference of the stochastic block model, which will be explained in Sec.~\ref{sec:SBM}. 
This section considers a spectral ordering method using the Bethe Hessian. 

The derivation of spectral ordering with the Bethe Hessian is analogous to that with the normalized Laplacian. 
However, instead of imposing an ellipsoidal constraint (\ref{EllipsoidalConstraint}), we introduce $\sum_{i=1}^{N} d_{i} x^{2}_{i}$ as a penalty term. 
Thus, we consider the following objective function:
\begin{align}
\frac{1}{2} \sum_{i,j} A_{ij} (x_{i} - x_{j})^{2} + \tau \, \sum_{i=1}^{N} d_{i} x^{2}_{i}, \label{BetheHessianH1}
\end{align}
where $\tau$ is an arbitrary constant (hyperparameter) that can be either positive or negative. 
To avoid the trivial minimizer $\vect{x} = \vect{0}$, we impose the spherical constraint $\sum_{i=1}^{N} x^{2}_{i} = 1$. 

Using Lagrange multipliers, the objective function to be minimized is then 
\begin{align}
& \frac{1}{2} \sum_{i,j} A_{ij} (x_{i} - x_{j})^{2} + \tau \sum_{i} d_{i} x^{2}_{i} 
-\lambda \left( \sum_{i=1}^{N} x_{i}^{2} - 1 \right) \notag\\
&= (1+\tau) \sum_{i,j} x_{i} B_{ij} x_{j}
-\lambda \left( \sum_{i=1}^{N} x_{i}^{2} - 1 \right), \label{BetheHessianH2}
\end{align}
where 
\begin{align}
B_{ij} = D_{ij} - r A_{ij} 
\hspace{20pt} 
\left( r = \frac{1}{1+\tau} \right)
\label{BetheHessian}
\end{align}
is a matrix element of Bethe Hessian. 
The extremum conditions in Eq.~(\ref{BetheHessianH2}) yield an eigenvalue equation with respect to $\mat{B}$. 

We estimate the optimal sequence $\hat{\vect{\pi}}$ as follows: 
\begin{align}
\hat{\vect{\pi}} = \left\{ \mathrm{rank}(\nu_{2i}) | i \in \mathcal{I} \right\}, \label{DiscretizationH2BetheHessian}
\end{align}
where $\vect{\nu}_{2}$ is the eigenvector associated with the second-smallest eigenvalue $\lambda_{2}$, i.e., 
\begin{align}
\mat{B} \, \vect{\nu}_{2} = \lambda_{2} \vect{\nu}_{2}. \label{BetheHessianEigenvalueEquation}
\end{align}
Note that there is no guarantee that $\vect{\nu}_{2}$ always provides the best estimate in terms of $H_{2}(\hat{\vect{\pi}}; \mat{A})$ among all the eigenvectors. 
In fact, we confirmed that the eigenvector that yields the best estimate in terms of $H_{2}(\hat{\vect{\pi}}; \mat{A})$ (when we employ the rounding rule in Eq.~(\ref{DiscretizationH2BetheHessian})) depends sensitively on the value of $r$, particularly when $r$ is small (see Sec.~\ref{sec:BetheHessianHyperparameter} in {\SupplementaryMaterials} for details). 
However, we employ Eq.~(\ref{DiscretizationH2BetheHessian}) because the estimate with $\vect{\nu}_{2}$ offers the smallest value of $H_{2}(\hat{\vect{\pi}}; \mat{A})$ as long as $r$ is sufficiently large. 

Throughout this study, we set $r = \sqrt{ \sum_{i} d^{2}_{i}/\sum_{i} d_{i} - 1}$ ($>1$), because it is a commonly employed value in spectral clustering. 
The hyperparameter $\tau$ is negative when $r>1$. 
Thus, $\{x_i\}$ for hub vertices are aligned near the ends of the real line $[-1,1]$ so that a sequence achieves a lower value of Eq.~(\ref{BetheHessianH1}) using the penalty term. 
By contrast, when $r<1$ ($\tau > 0$), $\{x_i\}$ for hub vertices are likely to be located near the origin, implying that the resulting sequence is similar to that obtained by the spectral ordering method based on the normalized Laplacian.

\subsection{Regularized Laplacian \label{sec:RegularizedLaplacian}}
During the past decade, it has been found that the performance of the Laplacian-based spectral clustering can be considerably improved by adding a constant value to every element in the adjacency matrix \cite{Amini2013,Joseph2016} or the diagonal elements in the degree matrix \cite{Chaudhuri2012,QinRohe2013}. 
Although the two variants of the Laplacian are often termed differently, we collectively refer to them as the regularized Laplacian \cite{Joseph2016} for simplicity, and we denote the former version of the regularized Laplacian as $\mat{\mathsf{L}}^{(\tau)}$ and the latter version as $\mat{\mathsf{L}}$. 
The spectral clustering method based on $\mathsf{L}$ can also be interpreted as a continuous relaxation of the minimization of the \textit{core cut} function \cite{ZhangRohe2018}. 
This section considers the spectral ordering method using a regularized Laplacian. 
 
Similar to the formulation of the spectral ordering method with the modularity matrix, we consider the continuous relaxation of $H_{2}\left( \vect{\pi}; \mat{A} \right)$ with a penalty term. 
We consider the following objective function: 
\begin{align}
\frac{1}{2} \sum_{i,j} A_{ij} (x_{i} - x_{j})^{2} + \tau N \, \mathrm{Var}\left[ \vect{x} \right], \label{RegularizedLaplacianH1}
\end{align}
which is minimized with respect to the continuous vector $\vect{x}$. $\tau$ is an arbitrary positive constant (hyperparameter) and 
\begin{align}
\mathrm{Var}\left[ \vect{x} \right] = \left( \frac{1}{N}\sum_{i=1}^{N} x^{2}_{i} - \left(\frac{1}{N}\sum_{i=1}^{N} x_{i}\right)^{2} \right)
\end{align}
is the variance with respect to the elements in $\vect{x}$. 
To ensure that $\vect{x}$ is not a vector of zeros, we impose the following ellipsoidal constraint: 
\begin{align}
\sum_{i=1}^N \left( d_{i} + \tau \right) x_i^{2} = 1. \label{EllipsoidalConstraintRegularizedLaplacian}
\end{align}
By incorporating this constraint, the objective function to be minimized is given by 
\begin{align}
& \frac{1}{2} \sum_{i,j} A_{ij} (x_{i} - x_{j})^{2} + \tau \sum_{i=1}^{N} x^{2}_{i} - \frac{\tau}{N}\left(\sum_{i=1}^{N} x_{i}\right)^{2} \notag\\
&\hspace{80pt} -\lambda \left( \sum_{i=1}^N \left( d_{i} + \tau \right) x_i^{2} - 1 \right). \label{RegularizedLaplacianH2}
%-\mu \sum_{i=1}^{N} x_{i}. 
\end{align}

Because a vector proportional to $\vect{1}$ is a trivial minimizer of Eq.~(\ref{RegularizedLaplacianH2}), we also impose the constraint that $\vect{x}$ is perpendicular to $\vect{1}$, i.e., $\sum_{i=1}^{N} x_{i} = 0$. 
Then, the extremum conditions in Eq.~(\ref{RegularizedLaplacianH2}) yield 
\begin{align}
\mat{A}^{(\tau)} \, \vect{\nu}_{2} = (1-\lambda_{2}) \left(\mat{D}+\tau \mat{I}\right) \vect{\nu}_{2}, 
\hspace{20pt} \left( A^{(\tau)}_{ij} = A_{ij} + \frac{\tau}{N} \right), \label{RegularizedLaplacianEigenvalueEquation1}
\end{align}
where $\mat{I}$ is the identity matrix. 
$\lambda_{2}$ is the second-smallest eigenvalue of the generalized eigenvalue equation and $\vect{\nu}_{2}$ is the associated generalized eigenvector. 
Equation (\ref{RegularizedLaplacianEigenvalueEquation1}) is equivalent to 
\begin{align}
\mat{\mathsf{L}}^{(\tau)} \, \vect{z}_{2} = \lambda_{2} \vect{z}_{2}, \label{RegularizedLaplacianEigenvalueEquation2}
\end{align}
where 
\begin{align}
\mat{\mathsf{L}}^{(\tau)} = \mat{I} - \left(\mat{D}+\tau \mat{I}\right)^{-1/2} \mat{A}^{(\tau)} \left(\mat{D}+\tau \mat{I}\right)^{-1/2} 
\end{align}
is the regularized Laplacian and $\vect{z}_{2} = \left(\mat{D}+\tau \mat{I}\right)^{1/2} \vect{\nu}_{2}$. 
Similar to the spectral ordering method with the normalized Laplacian, we estimate the optimal sequence $\hat{\vect{\pi}}$ as 
\begin{align}
\hat{\vect{\pi}} = \left\{ \mathrm{rank}\left((d_{i}+\tau)^{-1/2}  z_{2i} \right) | i \in \mathcal{I} \right\}. \label{DiscretizationH2NormalizedLaplacian}
\end{align}

\begin{figure}[t!]
  \centering
  \includegraphics[width= 0.99\columnwidth]{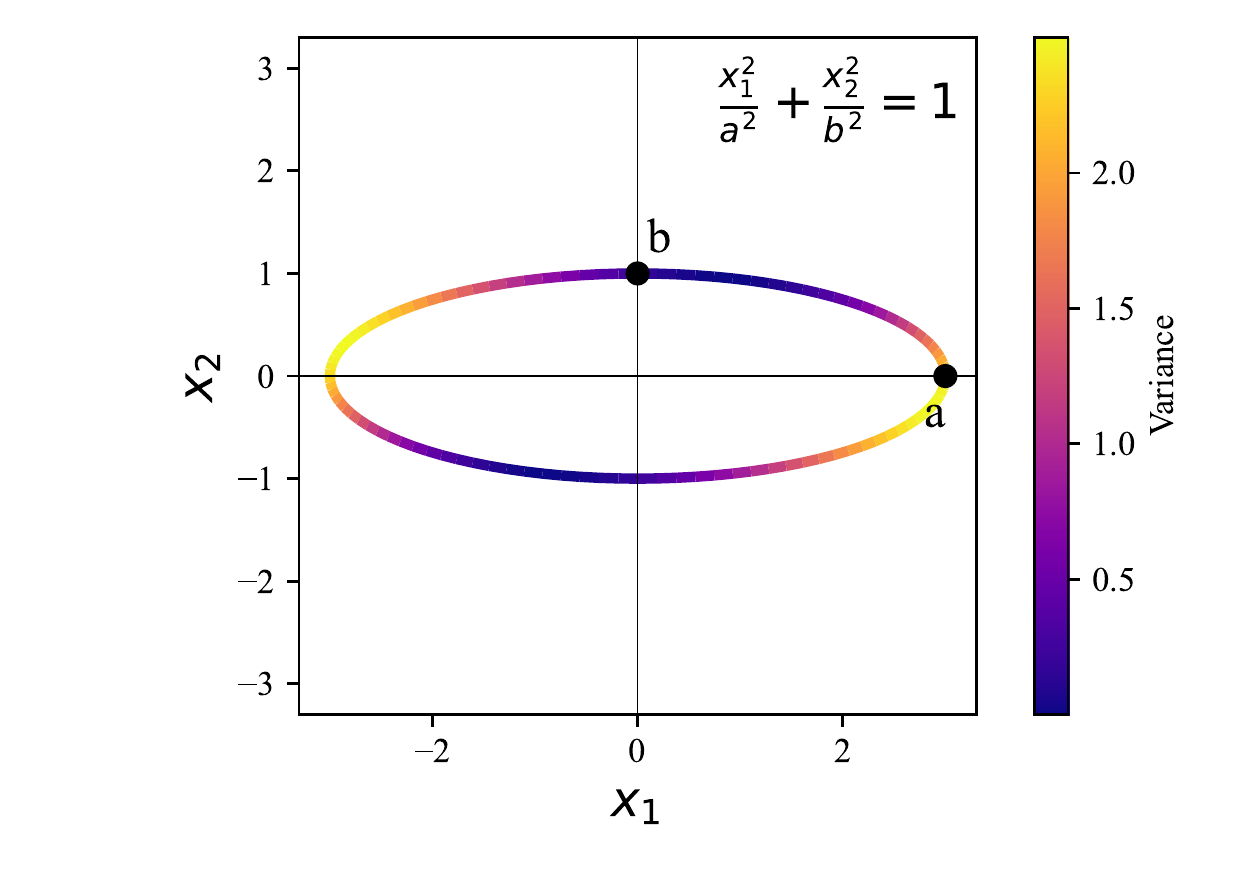}
  \caption{
  Ellipse equation $x^{2}_{1}/a^{2} + x^{2}_{2}/b^{2} = 1$ ($a=3$ and $b=1$), where $x_{2}$ corresponds to the variable for a hub vertex. 
  The color depth represents the variance $\mathrm{Var}\left[ \vect{x} \right]$ for $\vect{x} = (x_{1}, x_{2})$. 
  Although most of the coordinates on the ellipse have $x_{1} > x_{2}$, the variance is smaller when $x_{1}$ and $x_{2}$ are closer. 
  	}
  \label{fig:EllipseVariance}
\end{figure}

The contribution of hub vertices is more complicated than that of other methods. 
As shown in Fig.~\ref{fig:EllipseVariance}, whereas $\{x_{i}\}$ for the hub vertices tend to be relatively small because of the ellipsoidal constraint, the variance $\mathrm{Var}\left[ \vect{x} \right]$ is minimized when all $\{x_{i}\}$ have the same value.
Therefore, when the hyperparameter $\tau$ is small, the result is similar to that obtained by using the spectral ordering method based on the normalized Laplacian. 
As $\tau$ increases, the hub vertices are less likely to be located in the middle of the sequence because of the penalty term.

As mentioned above, we also consider 
\begin{align}
\mat{\mathsf{L}} = \left(\mat{D}+\tau \mat{I}\right)^{-1/2} \mat{A} \left(\mat{D}+\tau \mat{I}\right)^{-1/2} 
\end{align}
as the definition of a regularized Laplacian. 
Unlike $\mat{\mathsf{L}}^{(\tau)}$, only the degree matrix is perturbed by a constant value in $\mat{\mathsf{L}}$. 
If we consider 
\begin{align}
& \frac{1}{2} \sum_{i,j} A_{ij} (x_{i} - x_{j})^{2} + \tau \sum_{i=1}^{N} x^{2}_{i}. \label{RegularizedLaplacianH3}
\end{align}
as the objective function to be minimized, and impose the ellipsoidal constraint (\ref{EllipsoidalConstraintRegularizedLaplacian}), we obtain the eigenvalue equation with respect to $\mat{\mathsf{L}}$ as a result of the extremum conditions. 

If we also impose the constraint $\sum_{i=1}^{N} x_{i} = 0$ in Eq.~(\ref{RegularizedLaplacianH3}), this objective function becomes equivalent to Eq.~(\ref{RegularizedLaplacianH1}). 
We do not have such a constraint because $\mat{\mathsf{L}}$ does not have $\vect{1}$ as a trivial eigenvector unlike $\mat{\mathsf{L}}^{(\tau)}$. 
The eigenvectors of $\mat{\mathsf{L}}^{(\tau)}$ and $\mat{\mathsf{L}}$ are therefore distinct. 
As a spectral ordering method with the regularized Laplacian $\mat{\mathsf{L}}$, we replace $\vect{z}_{2}$ in Eq.~(\ref{DiscretizationH2NormalizedLaplacian}) with the eigenvector associated with the second-smallest eigenvalue of $\mat{\mathsf{L}}$.  
Here, we use the second-smallest value because $\mat{\mathsf{L}}$ approaches $\mat{\mathcal{L}}$ as $\tau\to0$, and $\mat{\mathcal{L}}$ has $\vect{z}_{1} \propto \vect{1}$.
%Hereafter, we employ this version because it is more computationally efficient. 
Hereafter, when we refer to the spectral ordering method with the regularized Laplacian, we employ $\mat{\mathsf{L}}$ because it is more computationally efficient. 
Throughout this study, we set $\tau$ as the average degree of the graph, as it is a commonly employed value \cite{QinRohe2013}. 

\begin{table*}[tb]
\caption{
Summary of constraints and penalty terms in spectral methods, and their effect on hub location in vertex sequence. 
}
\label{SummaryTableSpectralOrdering}
\begin{tabular}{lccl}
\hline
\textbf{Matrix}                                 & \textbf{Constraints}                                                                                                               & \textbf{Penalty terms}                                                                & \textbf{Effects on hub locations}                                                                                           \\ \hline
Unnormalized Laplacian \\ $\mat{L} = \mat{D} - \mat{A}$                & \begin{tabular}[c]{@{}c@{}}$\sum_{i=1}^{N} x^{2}_{i} = 1$\\ $\sum_{i=1}^{N} x_{i} = 0$\end{tabular}                                & -                                                                               & -                                                                                                       \\ \hline
Normalized Laplacian \\ $\mat{\mathcal{L}} = \mat{D}^{-\frac{1}{2}} \mat{L} \mat{D}^{-\frac{1}{2}}$        & \begin{tabular}[c]{@{}c@{}}$\sum_{i=1}^{N} d_{i} x^{2}_{i} = 1$\\ $\sum_{i=1}^{N} x_{i} = 0$\end{tabular}                          & -                                                                               & \begin{tabular}[c]{@{}l@{}}Concentrate around the middle\end{tabular}      \\ \hline
Modularity matrix \\ $\mat{Q} = \mat{A} + \frac{\vect{d}^{\top} \vect{d}}{2M}$ \\ $\left( \vect{d} = (d_{1}, \dots, d_{N}) \right)$                     & $\sum_{i=1}^{N} x^{2}_{i} = 1$                                                                                                     & $\frac{\left( \sum_{i} d_{i} x_{i} \right)^{2}}{2M} - \sum_{i} d_{i} x^{2}_{i}$ & \begin{tabular}[c]{@{}l@{}}Distribute at both ends\end{tabular} \\ \hline
Bethe Hessian \\ $\mat{B} = \mat{D} - r\mat{A}$                         & \begin{tabular}[c]{@{}c@{}}$\sum_{i=1}^{N} x^{2}_{i} = 1$\\ perpendicular to $\vect{\nu}_{1}$\end{tabular}                                                                                                     & $\tau \, \sum_{i=1}^{N} d_{i} x^{2}_{i}$                                        & \begin{tabular}[c]{@{}l@{}}$\tau > 0$: Concentrate around the middle \\ $\tau < 0$: Distribute at either/both ends \end{tabular}      \\ \hline
Regularized Laplacian \\ $\mat{\mathsf{L}}^{(\tau)} = \mat{I} - \mat{D}_{(\tau)}^{-1/2} \mat{A}^{(\tau)} \mat{D}_{(\tau)}^{-1/2}$ \\ $\left(\mat{D}_{(\tau)} = \mat{D}+\tau \mat{I}\right)$ & \begin{tabular}[c]{@{}c@{}}$\sum_{i=1}^N \left( d_{i} + \tau \right) x_i^{2} = 1$\\ $\sum_{i=1}^{N} x_{i} = 0$\end{tabular}        & $\tau N \, \mathrm{Var}\left[ \vect{x} \right]$                                 & \begin{tabular}[c]{@{}l@{}} small $\tau$: Concentrate around the middle \\ large $\tau$: Avoid  concentration around the middle \end{tabular}      \\ \hline
Regularized Laplacian \\ $\mat{\mathsf{L}} = \mat{D}_{(\tau)}^{-1/2} \mat{A} \mat{D}_{(\tau)}^{-1/2}$        & \begin{tabular}[c]{@{}c@{}}$\sum_{i=1}^N \left( d_{i} + \tau \right) x_i^{2} = 1$\\ perpendicular to $\vect{\nu}_{1}$\end{tabular} & $\tau \, \sum_{i=1}^{N} x^{2}_{i}$                                              & \begin{tabular}[c]{@{}l@{}} small $\tau$: Concentrate around the middle \\ large $\tau$: Avoid concentration around the middle \end{tabular}      \\ \hline
\end{tabular}
\end{table*}

The constraints and penalty terms for each method are summarized in Table~\ref{SummaryTableSpectralOrdering}. 
Compared to the classical method based on the normalized Laplacian, where hub vertices are concentrated around the middle of the sequence (``hub-centered''), the spectral ordering methods obtained with the modularity matrix, Bethe Hessian, and the regularized Laplacian may assign hub vertices at both ends of the sequence (``hub-at-the-corner''). 
Particularly, for the Bethe Hessian and the regularized Laplacian, we can choose the ``hub-centered'' or ``hub-at-the-corner'' alignment by tuning the hyperparameter. 

Although we found the penalties and constraints that provide the spectral ordering methods corresponding to the ones considered in spectral clustering, we have not confirmed whether these choices of penalties and constraints are unique. 
In addition, there is no guarantee that the resulting spectral ordering methods exhibit high performance in practice. 
The next section investigates the practical performance of these spectral ordering and clustering methods using synthetic and real-world datasets. 
Although one might expect all the methods to work similarly when the graph is close to regular, it is not trivial to determine whether this always holds; this is investigated using synthetic datasets in Secs.~\ref{sec:SBM} and \ref{sec:ORGM}. 
%\del{Note also that, even when the two matrices considered in the spectral ordering method are identical, the resulting vertex sequence can be different; this is because a tie-breaking is required when an eigenvector has multiple elements with the same value and the tie-breaking rule may not coincide depending on the implementation.}
%\teru{This is true, but maybe too much explanation about details..} 
The effect of heterogeneous degree distribution in each method is examined using real-world datasets in Sec.~\ref{sec:Realworld}.

%\section{Performance of matrix reordering methods for adjacency matrices}
\section{Performance analysis}\label{sec:PerformanceAnalysis}
We conduct a numerical performance analysis of the spectral ordering and clustering methods using synthetic graphs and real-world networks. 
For experiments on synthetic graphs, we consider a random graph model with a prespecified module structure, which is referred to as the stochastic block model ({\SBM}) \cite{holland1983stochastic,WangWong87,Peixoto2012}, and 
a random graph model with a prespecified sequentially local structure, which is referred to as the ordered random graph model (ORGM) \cite{KawamotoKobayashi2021}.

\subsection{Stochastic block model}\label{sec:SBM}
The {\SBM} is often used as a generative model for the inference of module structures in graphs \cite{Goldenberg2010,Peixoto2017tutorial} and in several theoretical studies in the community detection literature \cite{AbbleReview2017}. 
In the {\SBM}, each vertex has a ``planted'' (or preassigned) group assignment; we denote the corresponding partition as $\vect{\sigma}^{B}$.  Each vertex pair is connected by an edge, independently and randomly, based on the planted group assignments. 
The probabilities for the upper-right elements of the adjacency matrix are given as follows: 
\begin{align}
&\mathrm{Prob}\left[ \{ A_{ij} \}_{i < j} \right] = \prod_{i < j} p_{\sigma^{B}_{i} \sigma^{B}_{j}}^{A_{ij}} \left(1 - p_{\sigma^{B}_{i} \sigma^{B}_{j}} \right)^{1-A_{ij}}, \label{SBMProbability}
\end{align}
where $p_{k \ell}$ is the probability that a vertex in group $k$ and vertex in group $\ell$ are connected (in Eq.~(\ref{SBMProbability}), $k=\sigma^{B}_{i}$ and $\ell = \sigma^{B}_{j}$). 
We have $A_{ij} = A_{ji}$ for any pair of elements because we consider undirected graphs. 
In general, the {\SBM} can generate graphs with complex module structures. 
Herein, however, we focus on the {\SBM} with a community structure that is characterized by the following group-wise connection probability: 
\begin{align}
p_{k \ell} = 
\begin{cases}
p_{\mathrm{in}} & (k = \ell)\\
p_{\mathrm{out}} & (k \ne \ell)
\end{cases}, \label{DensityMatrixCommunity}
\end{align}
where $0 < p_{\mathrm{out}} \le p_{\mathrm{in}} \le 1$, that is, vertices are more densely connected within the same planted group than between different groups. 
In particular, when the group sizes are equal, it is common to parametrize the model using the average degree $c$ and the fuzziness parameter $\epsilon$, which are related to $p_{\mathrm{in}}$ and $p_{\mathrm{out}}$ as 
\begin{align}
c = \frac{N}{K} \left( p_{\mathrm{in}} + (K-1) p_{\mathrm{out}} \right), 
\hspace{20pt}
\epsilon = \frac{p_{\mathrm{out}}}{p_{\mathrm{in}}}.
\end{align}
As $\epsilon$ approaches unity, the planted community structure becomes less clear. 
This particular case of the {\SBM} is known as the planted partition model \cite{Condon2001}. 
For a given average degree $c$, the critical value of $\epsilon$ above which an algorithm cannot detect the planted block structure better than chance is called the (algorithmic) detectability limit \cite{Decelle2011,Mossel2015,Massoulie2014,Nadakuditi2012,Kawamoto2015Laplacian,KawamotoKabashimaEPL2015}. 

Using the {\SBM} and spectral ordering methods, we investigate the following questions: 
\begin{enumerate}
\item How would the reordered adjacency matrix look like? Can we visually identify the community structure through the matrix?
\item When and how would the spectral ordering methods lose their correlations with the planted partition in the {\SBM}? 
Are the spectral ordering methods superior or inferior to their clustering counterparts in detecting the planted partition? Does the choice of matrix matter?
\end{enumerate}
To answer these questions, we apply both spectral ordering and clustering methods to graphs generated by the {\SBM}.

\begin{figure}[tb]
  \centering
  \includegraphics[width= 0.9 \columnwidth]{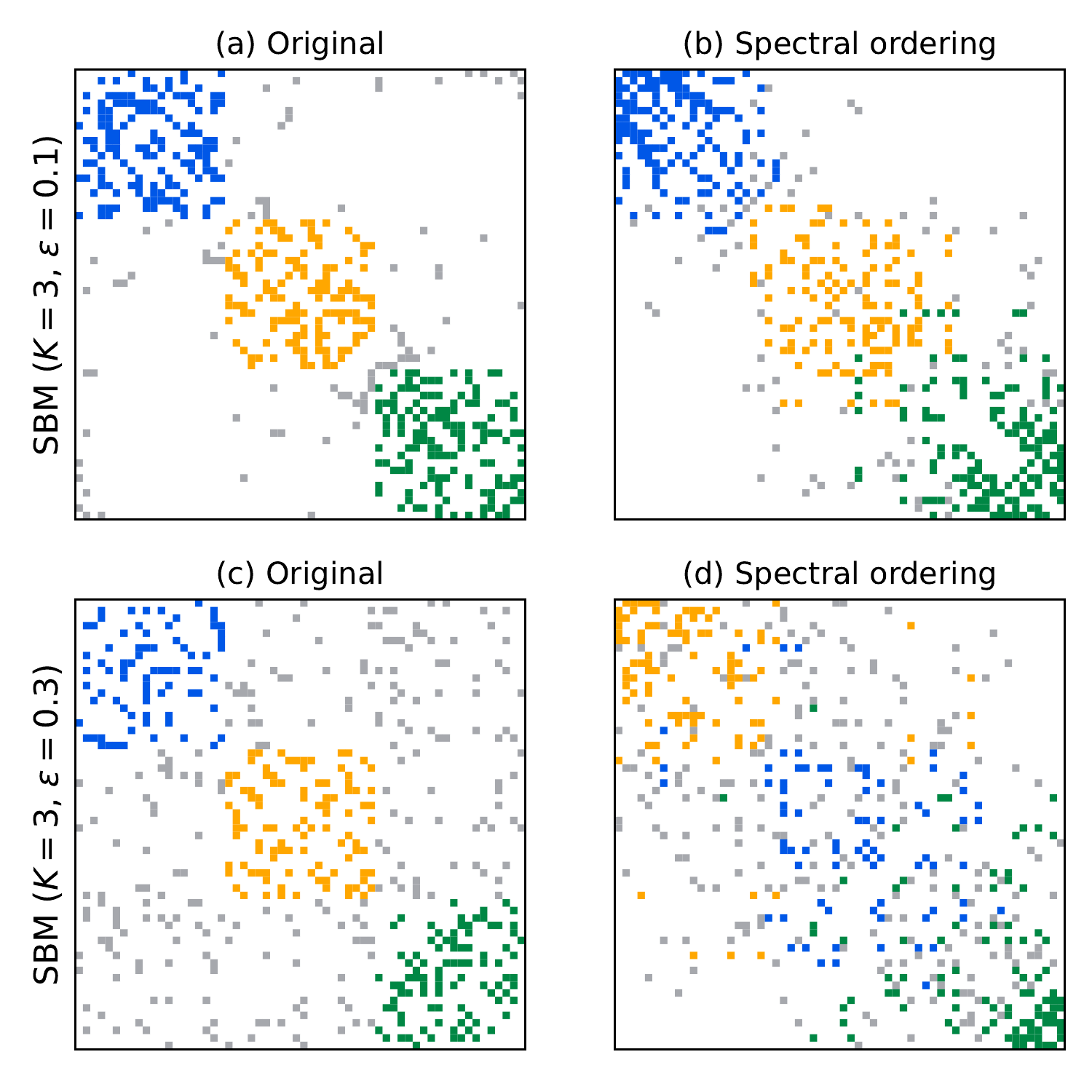}
  \caption{
  Spectral ordering with the regularized Laplacian for graphs generated by the {\SBM} ($N=60$ and $c=8$). 
  The top panels show the adjacency matrices of a graph with a strong community structure where vertices are aligned by (a) the sequence based on the planted partition $\vect{\sigma}^{B}$ and (b) an inferred sequence $\hat{\vect{\pi}}$. 
  The bottom panels show the adjacency matrices of a graph with a weak community structure where vertices are aligned based on (c) $\vect{\sigma}^{B}$ and (d) $\hat{\vect{\pi}}$.
  The matrix elements indicating the connections within the same planted group are represented in the same color; otherwise, the elements are colored in gray. 
	}
  \label{fig:SpectralClustering_SBM}
\end{figure}

We first investigate the former question. 
Figure \ref{fig:SpectralClustering_SBM} shows the results of spectral ordering applied to instances of {\SBM}. 
Vertices in the same planted group are indeed located closely in the inferred sequence when the community structure is strong. 
Even when the community structure is weak, the planted group labels and the inferred sequence are correlated. 
In both examples, the boundaries of the groups are ambiguous. 
Therefore, if we do not know the planted group labels (and without the coloring of the adjacency matrix elements), it is not clear whether the identified structure is a community structure or a banded structure from the reordered adjacency matrix. 
Note that, as discussed in \cite{KawamotoKobayashi2021}, even when we generate a graph from a uniformly random graph model, one could identify a weak banded structure owing to the ordering of vertices. 

Next, we address the latter question. 
The consistency between the sequence inferred by a spectral ordering method $\hat{\vect{\pi}}$ and planted partition $\vect{\sigma}^{B}$ is measured with the normalized {\LCE} $\Delta\left(\hat{\vect{\pi}}, \vect{\sigma}^{B} \right) / \overline{\Delta}\left( \{ N^{B}_{k} \} \right)$. 
Here, $\{ N^{B}_{k} \}$ is the set of group sizes in $\vect{\sigma}^{B}$ and $\overline{\Delta}\left( \{ N^{B}_{k} \} \right)$ is the {\LCE} under a random sequence defined in Eq.~(\ref{MeanLabelContinuityError}). 
When $\Delta\left( \hat{\vect{\pi}}, \vect{\sigma}^{B} \right)$ saturates (i.e., the normalized {\LCE} equals unity) as $\epsilon$ increases, the spectral ordering method does not infer $\vect{\sigma}^{B}$ better than random; it is deemed that the algorithm has reached the detectability limit.  

The consistency between the inferred partition $\hat{\vect{\sigma}}$ by a spectral clustering method and planted partition $\vect{\sigma}^{B}$ is measured using the normalized mutual information ({\NMI}) \cite{NMI}, which is defined as 
\begin{align}
\mathrm{NMI}\left( \hat{\vect{\sigma}}, \vect{\sigma}^{B} \right) 
= \frac{2 I\left(\hat{\vect{\sigma}}; \vect{\sigma}^{B}\right)}{H(\hat{\vect{\sigma}}) + H(\vect{\sigma}^{B})}, 
\end{align}
where 
\begin{align}
H(\vect{\sigma}) = - \sum_{k \in \{1, \dots, K\}} q(k) \log q(k) 
\hspace{20pt} \left( q(k) = \frac{N_{k}}{N} \right) \label{ShannonEntropy}
\end{align}
is the entropy with respect to the frequency of group labels, and
\begin{align}
I\left(\vect{\sigma}_{1}; \vect{\sigma}_{2}\right) = \sum_{k \in \{1, \dots, K\}} \sum_{k^{\prime} \in \{1, \dots, K\}} 
q(k, k^{\prime}) \log \frac{q(k, k^{\prime})}{q(k) q(k^{\prime})} \label{MutualInformation}
\end{align}
is the mutual information. 
Here, $q(k, k^{\prime})$ is the fraction of cooccurrences that a vertex belonging to group $k$ in partition $\vect{\sigma}_{1}$ belongs to group $k^{\prime}$ in partition $\vect{\sigma}_{2}$. 
The {\NMI} is unity when a pair of partitions coincides perfectly. %, while it is zero when the group assignments in two partitions do not coincide better than random. 
When $\mathrm{NMI}\left( \hat{\vect{\sigma}}, \vect{\sigma}^{B} \right)$ reaches (nearly) zero as $\epsilon$ increases, the spectral clustering method does not infer $\vect{\sigma}^{B}$ better than random, which again represents the detectability limit. 
The detectability analysis of spectral clustering methods is not new and has been analyzed in several theoretical and benchmark studies \cite{Nadakuditi2012,Kawamoto2015Laplacian,KawamotoKabashimaEPL2015,Darst2014,yang2016comparative}. 
We evaluate $\Delta\left(\hat{\vect{\pi}}, \vect{\sigma}^{B} \right)$ and $\mathrm{NMI}\left( \hat{\vect{\sigma}}, \vect{\sigma}^{B} \right)$ to compare the performances of the spectral ordering and clustering methods for each of the matrices considered in the previous section.

\begin{figure*}[tb]
  \centering
  \includegraphics[width= 2 \columnwidth]{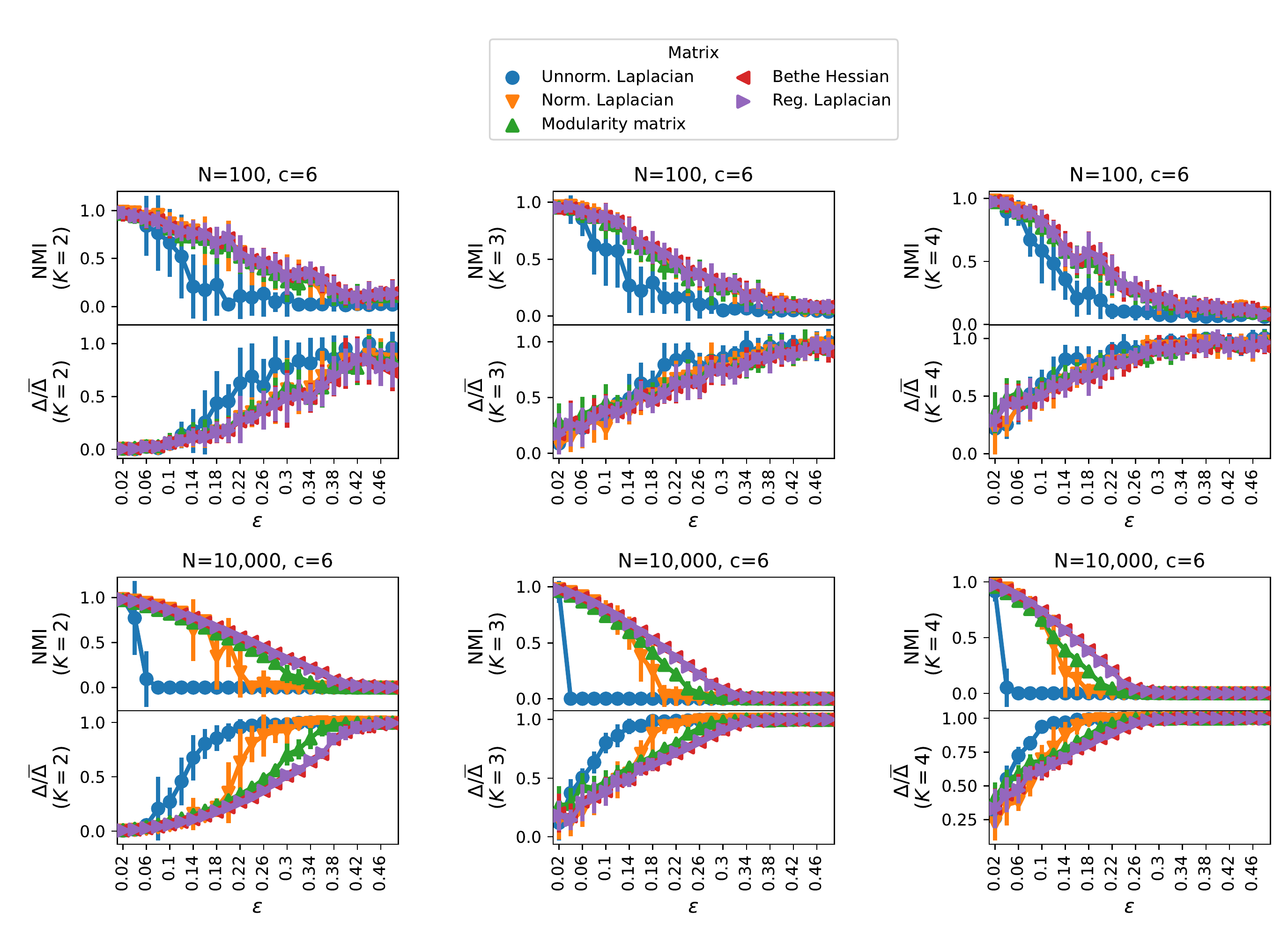}
  \caption{
  Detectability of the {\SBM} for the spectral ordering and spectral clustering methods. 
  The top panels show the result for small graphs with (a) $K=2$, (b) $K=3$, and (c) $K=4$. 
  The bottom panels show the result for large graphs with (d) $K=2$, (e) $K=3$, and (f) $K=4$. 
  In each panel, 
  the values of the {\NMI} obtained by the spectral clustering methods (top) and 
  the values of the {\LCE} obtained by the spectral ordering methods (bottom) are shown. 
  The horizontal axis represents the fuzziness of community structure $\epsilon$. 
  Each symbol and error bar represents the mean and the standard deviation of $30$ samples that are obtained with the same {\SBM} parameters. 
	}
  \label{fig:Detectability_SBM}
\end{figure*}

Figure \ref{fig:Detectability_SBM} shows the performances of the ordering and clustering methods based on the {\SBM} for different graph sizes $N$, numbers of blocks $K$, and fuzziness parameter $\epsilon$. 
When graphs are small (and thus relatively dense), there is no clear saturation in the curves of the {\LCE} and the {\NMI}, and it is difficult to evaluate whether the ordering methods or clustering methods exhibit superior performance in terms of the detectability limit. 
Moreover, the differences in performances are not noticeable among the different matrices, except for the unnormalized Laplacian. 
When graphs are large, we can clearly identify the saturation. 
For the unnormalized and normalized Laplacians, the values of {\LCE} gradually decrease, even when the values of {\NMI} saturate, indicating that the spectral ordering methods are superior to their clustering counterparts. 
By contrast, the detectability limits of the modularity matrix, regularized Laplacian, and Bethe Hessian are not very different between the ordering and clustering methods. 
In addition, the methods with the regularized Laplacian and Bethe Hessian perform similarly and are superior to the other matrices, whereas the methods with the unnormalized Laplacian are clearly inferior.

\subsection{Ordered random graph model}\label{sec:ORGM}
\begin{figure}[t!]
  \centering
  \includegraphics[width= \columnwidth]{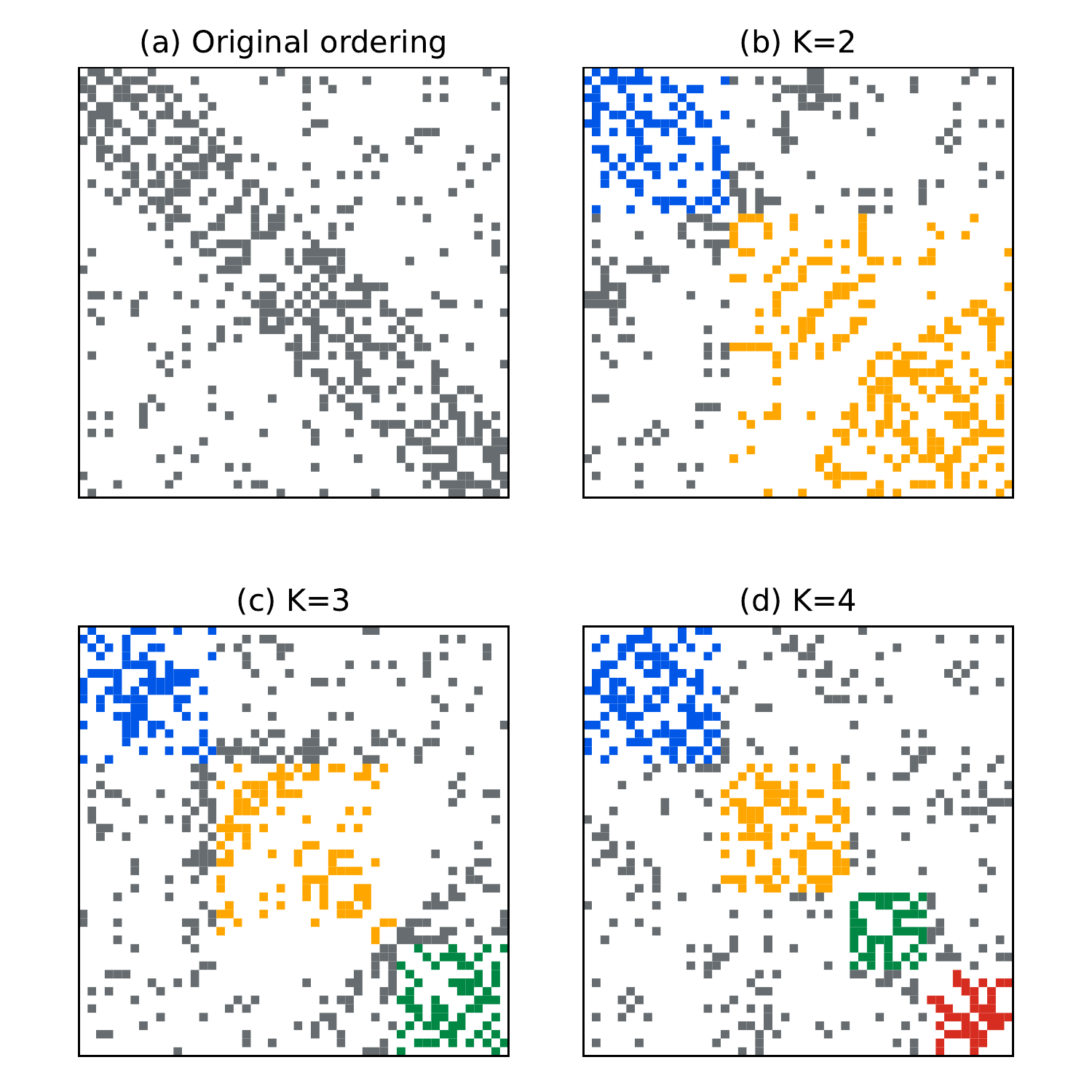}
  \caption{
  Results of the spectral clustering methods using the normalized Laplacian with different numbers of groups $K$, applied to a graph generated by the {\ORGM}. 
  The parameters of the {\ORGM} are $N=50$, $c=10$, $\epsilon=0.2$ and $r/N=0.16$. 
  (a) The vertices of the adjacency matrix are ordered based on the original ordering in the {\ORGM}.
  For panels (b)--(d), the vertices are ordered such that the vertices in the same inferred group are close to each other: (b) $K=2$, (c) $K=3$, and (d) $K=4$. 
  The nonzero matrix elements are represented by the same color when they are edges connecting the vertices within an inferred group; otherwise, the nonzero matrix elements are colored in gray. 
	}
  \label{fig:SpectralClustering_ORGM_Example}
\end{figure}

\begin{figure*}[ht!]
  \centering
  \includegraphics[width= 2 \columnwidth]{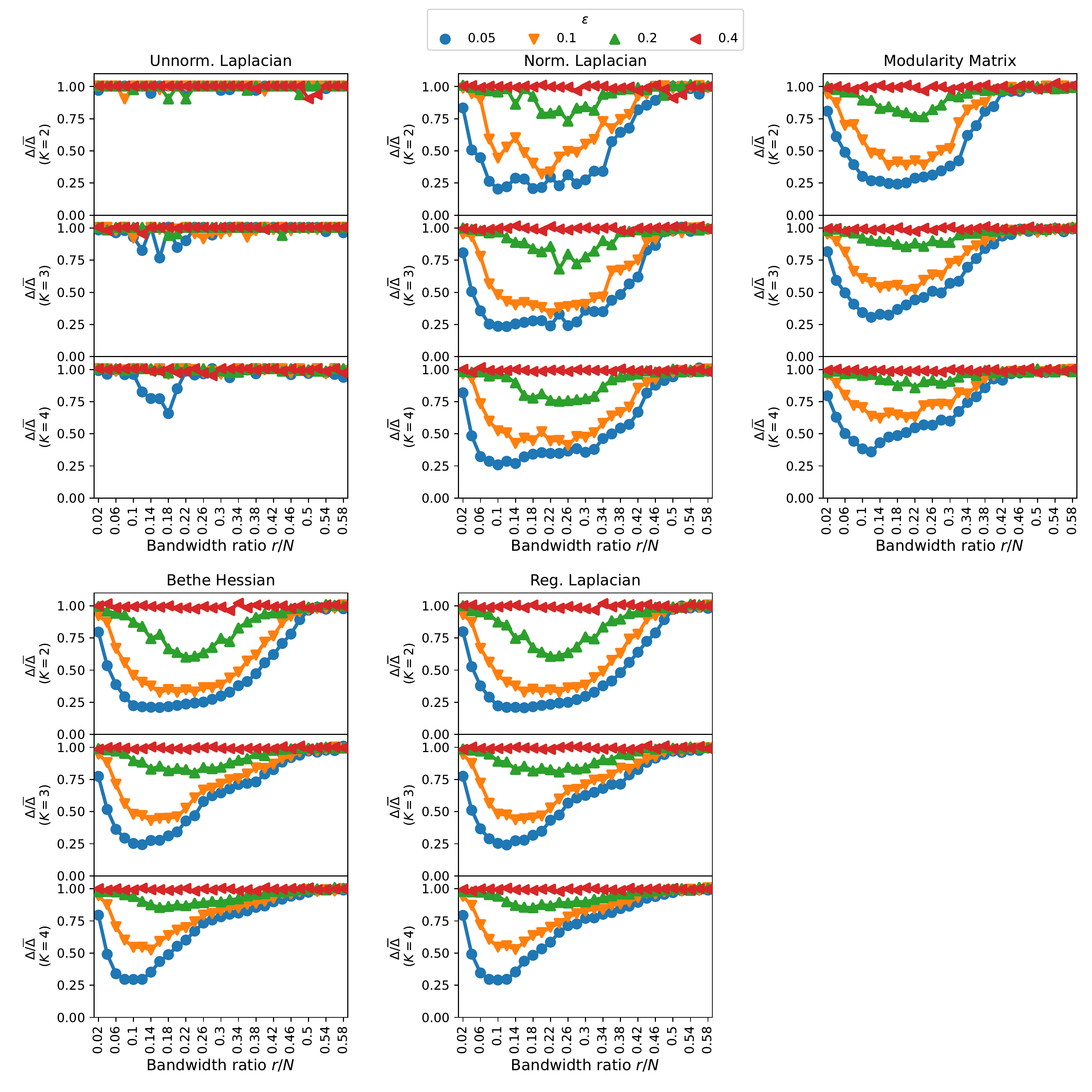}
  \caption{
  Performance of the spectral clustering method for the graphs generated by the {\ORGM}. 
  The graphs are generated by the {\ORGM} with $N=1,000$ and $c=6$. 
  Each panel shows the normalized {\LCE} $\Delta\left( \mathcal{I}, \hat{\vect{\sigma}} \right) / \overline{\Delta}(\{ \hat{N}_{k} \})$ for various parameter sets of the {\ORGM} for a matrix used in the spectral clustering method. 
  Each point represents the $10$-sample average of the normalized {\LCE} under the same parameter set. 
	}
  \label{fig:Detectability_ORGM}
\end{figure*}

We have observed how and to what extent the community structure can be inferred using spectral ordering methods. 
This section discusses the opposite scenario. 
That is, we analyze whether the spectral clustering methods can infer banded structures. 
To this end, we conduct a performance analysis using the {\ORGM}. 
This section uses the K-means method to determine the group labels in spectral clustering.

The vertex set in the {\ORGM} has a planted sequence, as the vertex set in the {\SBM} has a planted partition. 
We let the planted sequence coincide with the original sequence $\mathcal{I}$. 
The edges in the {\ORGM} are generated independently and randomly by referring to the planted sequence. 
We divide the space of the adjacency matrix elements into two regions, $\Omega_{\mathrm{in}}$ and $\Omega_{\mathrm{out}}$. 
$\Omega_{\mathrm{in}}$ (resp. $\Omega_{\mathrm{out}}$) is the set of elements in which an edge connects two vertices that are deemed to be ``close'' (resp. ``not close'') to each other. 
An edge is generated between a vertex pair with probability $p_{\mathrm{in}}$ if they are ``close'' and with probability $p_{\mathrm{out}}$ otherwise. 
Therefore, the probabilities of the upper-right elements of the adjacency matrix are given as follows: 
\begin{align}
& P \left( \{A_{ij}\}_{i<j} | \{ p_{ij} \} \right) 
= \prod_{i<j} p_{ij}^{A_{ij}} \left( 1 - p_{ij} \right)^{1-A_{ij}}, \\
& p_{ij} = 
\begin{cases}
p_{\mathrm{in}} & (i,j) \in \Omega_{\mathrm{in}} \\
p_{\mathrm{out}} & (i,j) \in \Omega_{\mathrm{out}} \\
\end{cases}.
\end{align}
We set the boundary of $\Omega_{\mathrm{in}}$ and $\Omega_{\mathrm{out}}$ as 
\begin{align}
\begin{cases}
\Omega_{\mathrm{in}} &= \{ (i,j) | |i-j| \le r \} \\
\Omega_{\mathrm{out}} &= \{ (i,j) | |i-j| > r \} 
\end{cases}, \label{ORGMboundary}
\end{align}
where $r$ is the bandwidth that specifies the boundary of the regions. 
Although Eq.~(\ref{ORGMboundary}) is a simple one, we note that the boundary in the {\ORGM} can be more complex in general. 
In the following, instead of $p_{\mathrm{in}}$ and $p_{\mathrm{out}}$, we specify the edge density by the average degree $c$ and the strength of the banded structure $\epsilon = p_{\mathrm{out}}/p_{\mathrm{in}}$; 
when $\epsilon = 0$, nonzero elements in the adjacency matrix are completely confined within $\Omega_{\mathrm{in}}$, whereas the model is uniformly random when $\epsilon = 1$. 
(See Fig.~\ref{fig:SpectralClustering_ORGM_Example}(a) for an example of the resulting adjacency matrix of the {\ORGM}). 
In summary, except for the number of vertices $N$, which is a nuisance parameter, the parameters in the {\ORGM} are the average degree $c$, strength of the banded structure $\epsilon$, and bandwidth ratio $r/N$. 
%: We denote $\mathrm{ORGM}(c,\epsilon,r/N)$ to represent the {\ORGM} with a specific parameter set. 

Using the {\ORGM} and the spectral clustering methods, we investigate the following questions: 
\begin{enumerate}
\item How would the reordered adjacency matrix look like? Can we visually identify the banded structure through the matrix?
\item How and when would the spectral clustering algorithms lose their correlations with the planted ordering in the {\ORGM}?
\end{enumerate}
%To answer these questions, we apply the spectral clustering methods to graphs generated by the {\ORGM}. 

We first investigate the former question. 
Figure \ref{fig:SpectralClustering_ORGM_Example} shows the results of a spectral clustering method with different values of $K$ applied to a graph generated by the {\ORGM}. 
A graph tends to be partitioned into equally-sized groups (see also Fig.~\ref{fig:MaxGroupSizeORGM} in the {\SupplementaryMaterials} for a quantitative evidence). 
Recall that we observe a banded structure through a spectral ordering method even when the graph is generated from the {\SBM} (Fig.~\ref{fig:SpectralClustering_SBM}). 
Analogously, we can identify block-diagonal structures in Fig.~\ref{fig:SpectralClustering_ORGM_Example} although the graph is generated from the {\ORGM}. 
This is an interesting observation because it implies that some of the community structures identified in the literature may be better described by banded structures. 

Figure \ref{fig:Detectability_ORGM} shows the normalized {\LCE} $\Delta\left( \mathcal{I}, \hat{\vect{\sigma}} \right) / \overline{\Delta}( \{ \hat{N}_{k} \} )$ between the planted sequence $\mathcal{I}$ and inferred partition $\hat{\vect{\sigma}}$, where $\{ \hat{N}_{k} \}$ is the set of group sizes in $\hat{\vect{\sigma}}$. 
The normalized {\LCE} is generally low when $r/N$ is not too small or large and $\epsilon$ is small. 

The existence of detectability limits is implied from Figure \ref{fig:Detectability_ORGM}. 
In the limit of $N \to \infty$, there exists a critical value of $\epsilon$ above which the normalized {\LCE} is unity for any value of $r/N$. 
Moreover, for a given $\epsilon$, there also exists an upper limit (and possibly a lower limit) of the bandwidth ratio $r/N$ beyond which a spectral clustering method is not correlated with the planted sequence better than a random guess. 
These critical values depend on the average degree (see Fig.~\ref{fig:phasediagramClusteringORGM} in the {\SupplementaryMaterials} for the numerical phase diagrams). 

Analogous to the analysis for the {\SBM}, the performance of the unnormalized Laplacian is notably inferior in terms of the normalized {\LCE}; in most of the parameter sets, it does not perform better than a random guess. 
The behaviors of the modularity matrix, Bethe Hessian, and regularized Laplacian are similar. 
Moreover, the results for the latter two matrices are apparently identical. 
In contrast to the analysis of graphs generated by the {\SBM}, the performance of the normalized Laplacian is as good as or even better than that of the Bethe Hessian and regularized Laplacian. 

%In addition to the value of the normalized {\LCE}, 
The inferior performance of the spectral clustering with the unnormalized Laplacian can also be characterized by the distribution of the group sizes $\{ N_{k} \}$. 
The fraction of the largest group $\max_{k} N_{k}/N$ is nearly unity, i.e., most of the vertices belong to the same group (see Fig.~\ref{fig:MaxGroupSizeORGM} in the {\SupplementaryMaterials} for the experimental results). 
In such a case, the result of clustering contains very little information about the inherent ordering in the graph; as shown in Fig.~\ref{fig:convexity}(b), the upper bound $\max \Delta$ and the mean value under the random sequence $\overline{\Delta}$ are small when a partition is highly skewed, reflecting the fact that the group labels tend to be aligned consecutively for any sequence. 
A possible mechanism for such skewed distributions of group sizes is the emergence of localized eigenvectors \cite{Kawamoto2015Laplacian,von2008consistency}, which deteriorates the performance of spectral clustering. 
However, we do not pursue the detailed mechanisms that could have caused the outcome obtained in this study. 

In summary, we have confirmed that some spectral clustering methods detect community structures that are correlated to the inherent sequential structure of the {\ORGM}, and that there are nontrivial limits of detectability.

\subsection{Real-world networks}\label{sec:Realworld}

\begin{figure*}[tb]
  \centering
  \includegraphics[width= 1.5\columnwidth]{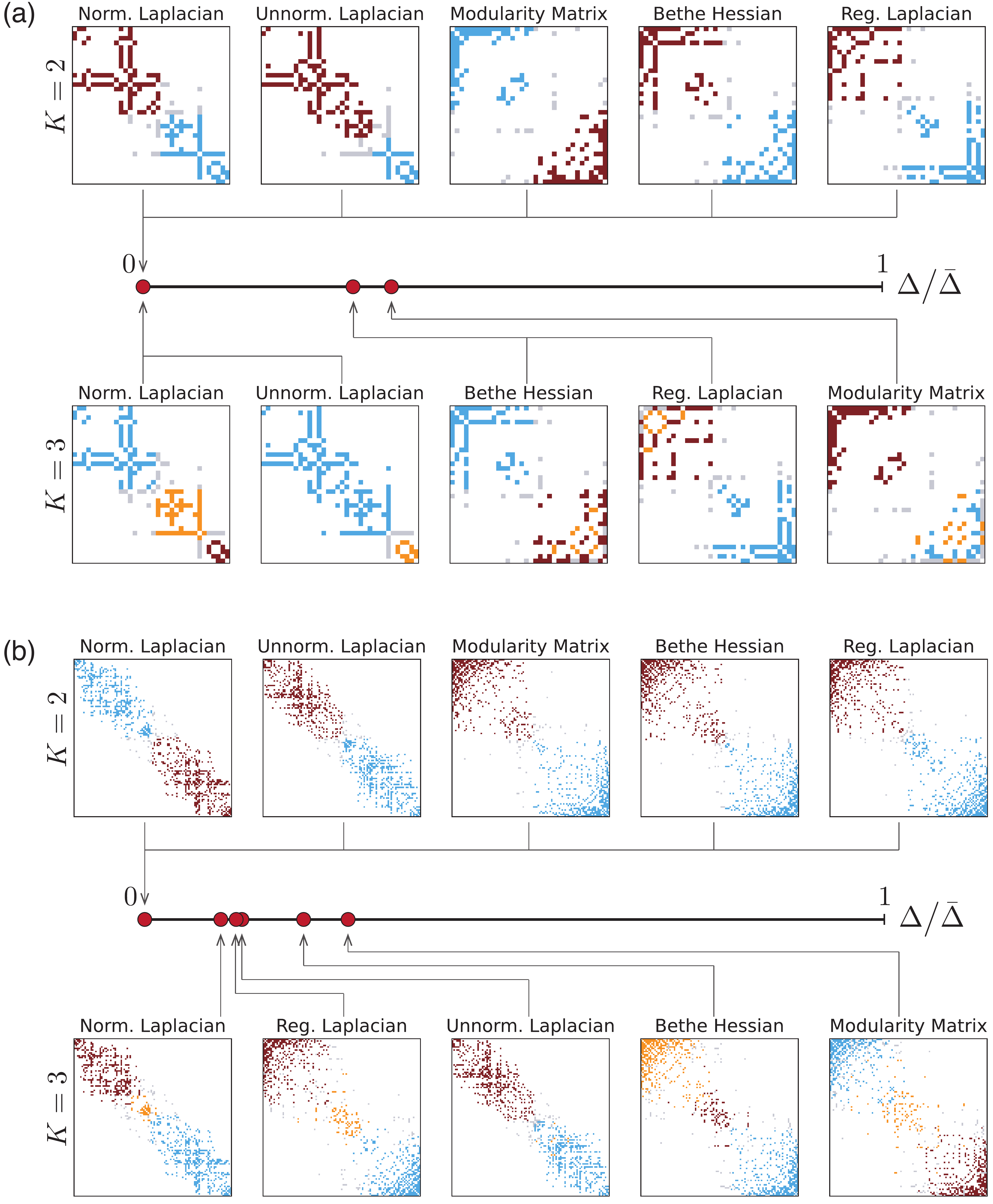}
  \caption{Adjacency matrix aligned with spectral ordering based on a matrix annotated at the top and its corresponding LCE, $\Delta/\bar{\Delta}$. 
         Colors denote vertex groups inferred by the K-means method. (a) Zachary's karate club network~\cite{zachary1977karate} and (b) a network of political books~\cite{polbook}.
	}
  \label{fig:empirical_adjacency}
\end{figure*}

We now apply the spectral ordering and clustering methods to five empirical adjacency matrices.
Descriptions of the empirical datasets examined are provided in Table~\ref{tab:dataset}. 
Note that many empirical datasets exhibit a high degree heterogeneity, whereas the synthetic graphs in Secs.~\ref{sec:SBM} and \ref{sec:ORGM} do not. 
As discussed in Sec.~\ref{sec:SpectralMethods}, spectral orderings with different matrices are characterized as the minimization problem of $H_2$ with different constraints and penalty terms (Table~\ref{SummaryTableSpectralOrdering}), and these differences become prominent when vertex degrees are heterogeneous. 

In Fig.~\ref{fig:empirical_adjacency}, we see a banded structure for the vertex orderings based on the normalized and unnormalized Laplacians, $\mat{L}$ and $\mat{\mathcal{L}}$, for the karate club (Fig.~\ref{fig:empirical_adjacency}(a)) and political books datasets (Fig.~\ref{fig:empirical_adjacency}(b)), where the hub vertices tend to be located around the middle of the optimized sequence. 
In contrast, the ordering method with the modularity matrix $\mat{Q}$ locates vertices with large degrees at both ends of the sequence, as expected from the penalty term in the objective function (\ref{ModularityPenalization}). 
A similar observation applies to the methods using the Bethe Hessian $\mat{B}$ and regularized Laplacian $\mat{\mathsf{L}}$ (Figs.~\ref{fig:empirical_adjacency}, \ref{fig:empirical_adjacency_K2} and \ref{fig:empirical_adjacency_K4}).
Importantly, however, vertex orderings based on these matrices is critically influenced by the regularization parameter $\tau$.

For many empirical graphs, the hyperparameter $r = \sqrt{ \sum_{i} d^{2}_{i}/\sum_{i} d_{i} - 1}$ in the Bethe Hessian takes a large positive value, i.e., $\tau<0$ (Eq.~(\ref{BetheHessianH1})). 
Thus, the penalty term $\tau\sum d_ix_i^2$ contributes to reducing the objective function. 
Hence, hub vertices tend to be aligned at the ends of the vertex sequence. % when implementing the spectral ordering method on empirical data.
However, the spectral ordering with the regularized Laplacian has an exogenous regularization parameter $\tau$ in its constraint and penalty terms (see Eqs.~(\ref{RegularizedLaplacianH2}) and (\ref{RegularizedLaplacianH3})), where we set $\tau$ as the average degree. 
As discussed in Sec.~\ref{sec:RegularizedLaplacian}, a larger value of $\tau$ tends to avoid locating hub vertices around the middle of the sequence. 
Although the validation analysis based on synthetic graphs suggested that the sequences inferred based on these matrices are fairly similar (Sec.~\ref{sec:PerformanceAnalysis}), they do not necessarily coincide in general. 
Note also that, as $\tau \to 0$, the Bethe Hessian $\mat{B}$ approaches the unnormalized Laplacian $\mat{L}$, and the regularized Laplacian $\mat{\mathsf{L}}$ approaches the normalized Laplacian $\mat{\mathcal{L}}$ (Table~\ref{SummaryTableSpectralOrdering}). 
Therefore, when $\tau$ is small in absolute value, the optimal vertex sequences based on the Bethe Hessian and regularized Laplacian are close to those obtained from the unnormalized and normalized Laplacians, respectively (Figs.~\ref{fig:karate_tau_K2} and \ref{fig:polbooks_tau_K2}). 
Indeed, the location of vertices with large degrees in optimal vertex sequence can be tuned by varying $\tau$, from a ``hub-centered'' alignment to a ``hub-at-the-corner'' alignment. 

Figure \ref{fig:empirical_adjacency} also shows the normalized {\LCE} representing the consistency between the inferred sequence $\hat{\vect{\pi}}$ and group labels $\hat{\vect{\sigma}}$ for each matrix used in the spectral ordering and clustering methods. 
When we set $K=2$ in the clustering method, as shown in Figs.~\ref{fig:empirical_adjacency}(a) and \ref{fig:empirical_adjacency}(b), $\hat{\vect{\pi}}$ and $\hat{\vect{\sigma}}$ are perfectly consistent in terms of the {\LCE}.
For $K\geq 3$, the {\LCE}s are mostly lower than $0.8$ and are typically approximately $0.5$ (Fig.~\ref{fig:empirical_adjacency} and \ref{fig:norm_LCE_K2}--\ref{fig:norm_LCE_K6}), suggesting that the optimized vertex sequences using spectral ordering convey some information about a non-random structure. 
%\textcolor{red}{The empirical {\LCE}s are thus comparable to those of the synthetic models (i.e., SBM and ORGM) that exhibit strong structures (Figs.~\ref{fig:Detectability_SBM} and \ref{fig:Detectability_ORGM}), suggesting that the optimized vertex sequences using spectral ordering convey some information about a non-random structure.}
We also find that some methods yield similar {\LCE}s for all datasets, whereas the {\LCE}s obtained with the (un)normalized Laplacian exhibit different behaviors (Figs.~\ref{fig:norm_LCE_K2}--\ref{fig:norm_LCE_K6}). 
This is consistent with the previous numerical observation that the spectral ordering based on the (un)normalized Laplacian is quite distinct from those obtained from the modularity matrix, Bethe Hessian, and regularized Laplacian (Fig.~\ref{fig:empirical_adjacency}).
%Another reason for having different {\LCE}s is that the clustering method (i.e., K-means) may not work equally well for all matrices. 
%We find that the sizes of inferred groups in empirical graphs can be extremely heterogeneous (e.g., there are only a few vertices in a group), in which case the {\LCE} can be fairly small. 

Interestingly, despite having distinct optimized sequences using different objective functions, the value of the normalized {\LCE} can be very close to each other. 
Therefore, adjacency matrices may exhibit the same or similar structures from the perspective of community structure, and they are differentiated only by detailed orderings within each group.

\section{Summary and discussion}\label{sec:Discussion}
This study analyzed the relationship between the ordering and clustering methods for graphs by quantifying the extent to which vertices close to each other in the optimized sequence have the same group label through the {\LCE}.  
To obtain analytical insight into spectral ordering, we first showed that the spectral ordering problem is formulated as a minimization of the squared sequential distance $H_{2}$ subject to a particular penalty function and constraints, depending on the matrix representation of a graph (i.e., normalized Laplacian, modularity matrix, etc). 
The numerical results suggested that the spectral ordering methods, except that based on unnormalized Laplacian, often yield optimized sequences such that vertices in the same group are close to each other; that is, the normalized {\LCE}s are considerably below $1$ as long as strong community structures exist. 

Several issues remain to be addressed in future studies.
First, we defined {\LCE} to quantify the continuity of group labels for a given vertex sequence.
The consistency between ordering and clustering can also be measured in other ways; for example, one can quantify the continuity of indices in a vertex sequence for given group labels on the vertices, whereas the {\LCE} quantifies the continuity of group labels for a given vertex sequence. 
 Second, we focused on unipartite graphs for which the connectivities are represented by square matrices (i.e., adjacency matrices).
 In principle, the proposed method can also be applied to study non-square matrices, such as bipartite graphs.
 Third, we implemented ordering and clustering methods independently and examined their consistency.
 Given that we found some consistency between the two, it would be possible to develop a clustering method that incorporates information about the inherent vertex sequence.
 Analogously, the spectral ordering method can be adjusted in such a way that the obtained vertex sequence reflects group labels.
 We expect our paper will stimulate further research in these directions.

% Specify following sections are appendices. Use \appendix* if there
% only one appendix.
\appendix

\section{Upper bound of the label continuity error}\label{sec:MaximumLCE}
\begin{figure}[t!]
  \centering
  \includegraphics[width= 0.9\columnwidth]{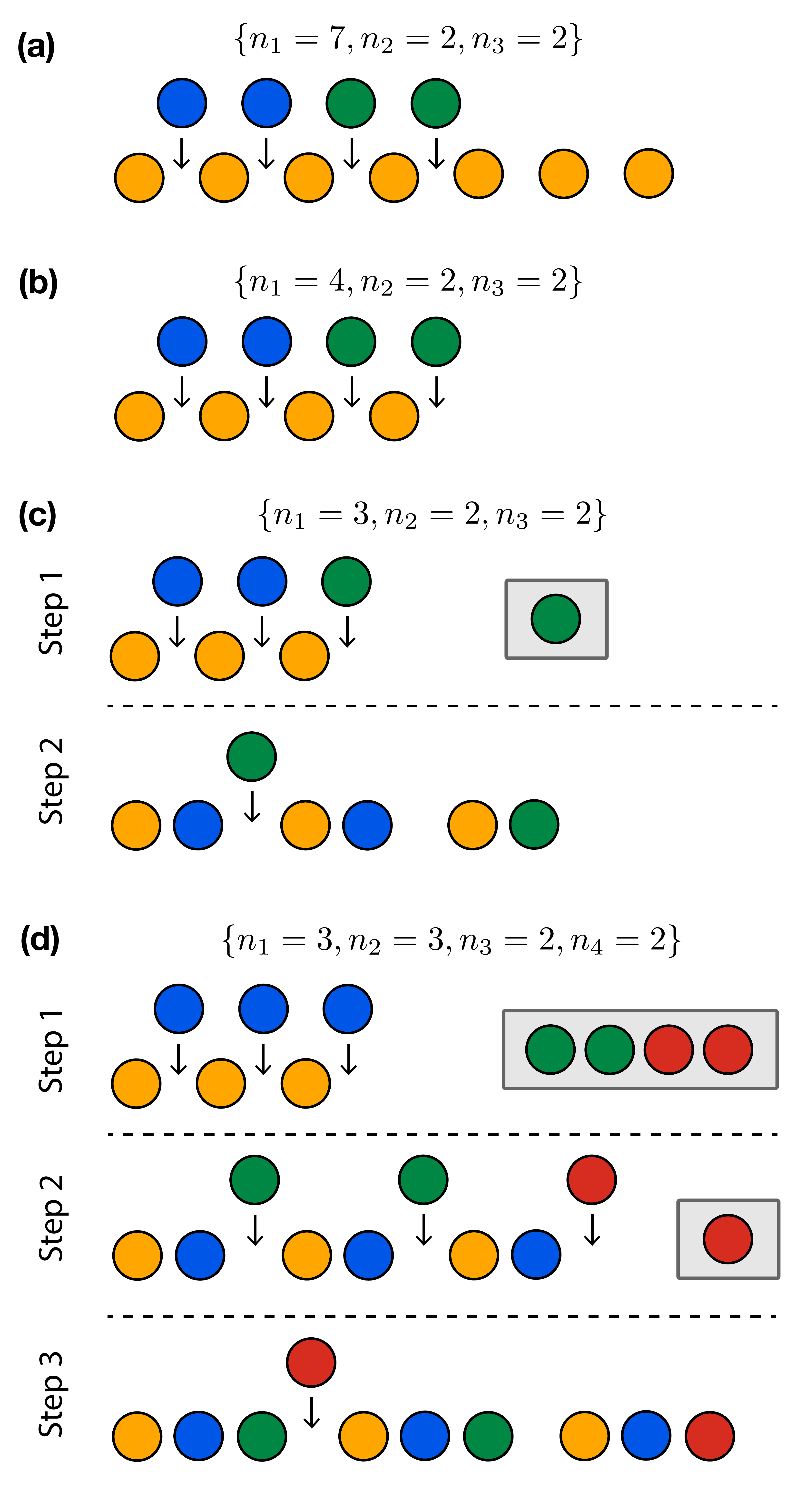}
  \caption{
  Vertex sequence yielding the maximum {\LCE} for a given group sizes $\{ N_{k} \}$. 
  Panels (a) and (b) are cases where $N_{1} > \left\lceil N/2 \right\rceil$ and $N_{1} = N/2$, respectively, and panels (c) and (d) are cases where $N_{1} < N/2$. 
  The sequence with the maximum {\LCE} can be constructed by aligning the vertices with different labels alternately in the procedure shown in each step. 
  Vertices in the box indicate that they are to be aligned in the following steps. 
  The vertex indices are omitted because they are not essential for the construction of a sequence. 
	}
  \label{fig:SchematicMaxLCE}
\end{figure}

We derive the upper bound of the {\LCE} by explicitly constructing a worst-case sequence. 
We assume that a partition $\vect{\sigma}$ is given (i.e., the number of groups $K$ and group sizes $\{ N_{k} \}$ are given), and the first group is the largest group (i.e., $\max_{k} N_{k} = N_{1} = |V_{1}|$). 
When $N_{1}$ satisfies $N_{1} > \left\lceil N/2 \right\rceil$, some vertices in $V_{1}$ must be aligned consecutively. 
As exemplified in Fig.~\ref{fig:SchematicMaxLCE}(a), the {\LCE} is maximized when the vertices in $V_{1}$ and those in $\cup_{k>1} V_{k}$ are aligned alternately as possible.
In this case, there are $2(N-N_{1})$ vertices that are aligned alternately with different group labels, and the label continuity is $\mathcal{C} = (2N_{1} -N -1)/(N-1)$. 
Therefore, the maximum {\LCE} leads to
\begin{align}
\Delta 
&= 1 - \frac{K-1}{N-1} - \frac{2N_{1} -N -1}{N-1} \notag\\
&= \frac{2 (N-N_{1})}{N-1} - \frac{K-1}{N-1}, 
\end{align}
which corresponds to the upper case of Eq.~(\ref{MaxLCE}). 

When $N_{1}$ is less than or equal to the sum of the vertices in all other groups (Figs.~\ref{fig:SchematicMaxLCE}(b), \ref{fig:SchematicMaxLCE}(c), and \ref{fig:SchematicMaxLCE}(d)), vertices can be aligned such that no group labels are consecutive. 
Such a sequence is constructed as follows. 
We first align the vertices in $V_{1}$ and the vertices in $\cup_{k>1} V_{k}$ as alternately as possible. 
In this step, all the vertices in $V_{1}$ are aligned, and there are $\sum_{k>1}N_{k} - N_{1}$ vertices that are not yet aligned; 
here, in $\cup_{k>1} V_{k}$, we preferentially consume the labels with larger $N_{k}$ (Fig.~\ref{fig:SchematicMaxLCE}(b) and Step 1 in Figs.~\ref{fig:SchematicMaxLCE}(c) and \ref{fig:SchematicMaxLCE}(d)). 
When there are remaining vertices, we regard a set of alternately-aligned vertices as a fundamental unit and treat all such sets as ``super vertices'' with the same labels. 
We then align the super vertices and the remaining vertices in the same manner as in the previous step. 
We repeat this procedure until all vertices are aligned. 
We can always align vertices and super vertices alternately because the number of remaining vertices with the same label never exceeds the number of already aligned vertices or super vertices. 
Therefore, we can establish a sequence for which the label continuity $\mathcal{C}$ is zero, and the upper bound of the {\LCE} leads to
\begin{align}
\Delta 
&= 1 - \frac{K-1}{N-1}.
\end{align}

%%%%%%%%%%%%%%%%%%%%%%%%%%%%%%%%%%%%%%%%%%%%%%%%%%%%%%%%%%%
%%%%%%%%%%%%%%%%%%%%%%%%%%%%%%%%%%%%%%%%%%%%%%%%%%%%%%%%%%%

\section{Variance of the label continuity error in random partitions}\label{sec:VarSER}
The second moment of $\Delta$ is 
\begin{align}
&\mathbb{E}\left[ \Delta^{2} \right] \notag\\
&= \sum_{\vect{\sigma}^{\ast}} P(\vect{\sigma}^{\ast}) 
\left( \frac{N-K}{N-1} - \frac{\sum_{i=1}^{N-1} \delta\left( \sigma^{\ast}(i), \sigma^{\ast}(i+1) \right)}{N-1} \right)^{2} \notag\\
&= \left( \frac{N-K}{N-1} \right)^{2} 
- 2 \frac{N-K}{N-1} \mathbb{E}\left[ \mathcal{C} \right] \notag\\
&+ \frac{1}{(N-1)^{2}} \sum_{i=1}^{N-1} \sum_{\sigma^{\ast}_{i}, \sigma^{\ast}_{i+1}} P(\sigma^{\ast}_{i}) P(\sigma^{\ast}_{i+1}) \delta\left( \sigma^{\ast}(i), \sigma^{\ast}(i+1) \right) \notag\\
&+ \frac{2}{(N-1)^{2}} \sum_{i=1}^{N-2} \sum_{\sigma^{\ast}_{i}, \sigma^{\ast}_{i+1}, \sigma^{\ast}_{i+2}} P(\sigma^{\ast}_{i}) P(\sigma^{\ast}_{i+1}) P(\sigma^{\ast}_{i+2}) \notag\\
&\hspace{50pt}\times \delta\left( \sigma^{\ast}(i), \sigma^{\ast}(i+1) \right) \delta\left( \sigma^{\ast}(i+1), \sigma^{\ast}(i+2) \right) \notag\\
&+ \frac{1}{(N-1)^{2}} \sum_{\substack{i,j \\ (|i-j| > 2)}} 
\sum_{\substack{\sigma^{\ast}_{i}, \sigma^{\ast}_{i+1},\\ \sigma^{\ast}_{j}, \sigma^{\ast}_{j+1}}} 
P(\sigma^{\ast}_{i}) P(\sigma^{\ast}_{i+1}) P(\sigma^{\ast}_{j}) P(\sigma^{\ast}_{j+1}) \notag\\
&\hspace{50pt}\times \delta\left( \sigma^{\ast}(i), \sigma^{\ast}(i+1) \right) 
\delta\left( \sigma^{\ast}(j), \sigma^{\ast}(j+1) \right) \\
&= \left( \frac{N-K}{N-1} \right)^{2} 
- 2 \frac{N-K}{N-1} \sum_{k=1}^{K} \left( \frac{N_{k}}{N} \right)^{2} 
+ \frac{1}{N-1} \sum_{k=1}^{K} \left( \frac{N_{k}}{N} \right)^{2} \notag\\
&+ \frac{2(N-2)}{(N-1)^{2}} \sum_{k=1}^{K} \left( \frac{N_{k}}{N} \right)^{3}
+ \frac{(N-2)(N-3)}{(N-1)^{2}} \left( \sum_{k=1}^{K} \left( \frac{N_{k}}{N} \right)^{2} \right)^{2}. \notag\\
\end{align}
Thus, the variance $\mathrm{Var}[\Delta]$ is 
\begin{align}
\mathrm{Var}[\Delta] 
&= \mathbb{E}\left[ \Delta^{2} \right] - \mathbb{E}\left[ \Delta \right]^{2} \notag\\
%&= \frac{1}{N-1} \sum_{k=1}^{K} \left( \frac{N_{k}}{N} \right)^{2} 
%+ \frac{2(N-2)}{(N-1)^{2}} \sum_{k=1}^{K} \left( \frac{N_{k}}{N} \right)^{3}
%+ \left( \frac{(N-2)(N-3)}{(N-1)^{2}} - 1 \right) \left( \sum_{k=1}^{K} \left( \frac{N_{k}}{N} \right)^{2} \right)^{2} \notag\\
&= \frac{1}{N-1} \sum_{k=1}^{K} \left( \frac{N_{k}}{N} \right)^{2} 
+ \frac{2(N-2)}{(N-1)^{2}} \sum_{k=1}^{K} \left( \frac{N_{k}}{N} \right)^{3} \notag\\
&\hspace{20pt}- \frac{3N - 5}{(N-1)^{2}} \left( \sum_{k=1}^{K} \left( \frac{N_{k}}{N} \right)^{2} \right)^{2}. 
\end{align}

%%%%%%%%%%%%%%%%%%%%%%%%%%%%%%%%%%%%%%%%%%%%%%%%%%%%%%%%%%%

\section{Probability distribution of the label continuity error in random partitions}\label{sec:NullDistribution}
This section derives the probability distribution of the label continuity error $\Delta\left( \vect{\pi}, \vect{\sigma} \right)$ when group labels are assigned randomly based on bootstrapped group labels $\vect{\sigma}^{\ast}$. 

To derive the probability distribution of $\Delta\left( \vect{\pi}, \vect{\sigma} \right)$, it is sufficient to calculate that of label continuity $\mathcal{C}\left( \vect{\pi}, \vect{\sigma} \right)$. 
The probability of $(N-1) \mathcal{C} = m$ is 
\begin{align}
& P\left[ (N-1) \mathcal{C} = m \right] \notag\\
&= \sum_{\vect{\sigma}^{\ast}} P(\vect{\sigma}^{\ast}) \delta\left( m, \sum_{i=1}^{N-1} \delta\left( \sigma^{\ast}(i), \sigma^{\ast}(i+1) \right) \right) \notag\\
&= \sum_{\vect{\sigma}^{\ast}} \prod_{i=1}^{N} \frac{N_{\sigma^{\ast}(i)}}{N} 
\oint \frac{dz}{2\pi i} z^{\sum_{i=1}^{N-1} \delta\left( \sigma^{\ast}(i), \sigma^{\ast}(i+1) \right) - m -1} \notag\\
&= \oint \frac{dz}{2\pi i} z^{-(1+m)} \sum_{\vect{\sigma}^{\ast}} 
\frac{N_{\sigma^{\ast}(N)}}{N} \prod_{i=1}^{N-1} \left( \frac{N_{\sigma^{\ast}(i)}}{N} z^{\delta\left( \sigma^{\ast}(i), \sigma^{\ast}(i+1) \right)} \right) \notag\\
&= \oint \frac{dz}{2\pi i} z^{-(1+m)} \vect{1}^{\top} \mat{D} \left(\mat{F} \mat{D}\right)^{N-1} \vect{1}, %\notag\\
%&= \frac{1}{N^{N}} \oint \frac{dz}{2\pi i} z^{-(1+m)} \sum_{k=1}^{K} \lambda_{k}^{N-1} \bra{1}\ket{v}_{k} \bra{v}_{k}\ket{n} 
\label{RandomSERProb1}
\end{align}
where 
\begin{align}
& \mat{D} \equiv \mathrm{diag}\left( \frac{N_{1}}{N}, \dots, \frac{N_{K}}{N} \right), 
%& \mat{F} \equiv \ket{1}\bra{1} + (z-1) \mat{I}. 
& \mat{F} \equiv \vect{1}\vect{1}^{\top} + (z-1) \mat{I}. 
\end{align}
Here, $\mat{I}$ is the identity matrix. 
In Eq.~(\ref{RandomSERProb1}), we used the identity 
\begin{align}
\delta\left( x,y \right) = \oint \frac{dz}{2\pi i} \frac{1}{z^{x-y+1}}, 
\end{align}
which is an integral around the origin of the complex plane. 

Using the eigenvalue decomposition, $\mat{F}$ can be expressed as 
\begin{align}
%\frac{z - 1 + K}{K} \ket{1} \bra{1} + (z-1) \sum_{k=2}^{K} \ket{u_{k}} \bra{u_{k}}, 
\frac{z - 1 + K}{K} \vect{1}\vect{1}^{\top} + (z-1) \sum_{k=2}^{K} \vect{u}_{k}\vect{u}_{k}^{\top}, 
\end{align}
where $\vect{u}_{k}$ ($2 \le k \le K$) is an eigenvector of $\mat{F}$ that is perpendicular to $\vect{1}$, and we have 
\begin{align}
%\mat{F} \mat{D} \ket{1} = \frac{z - 1 + K}{K} \ket{1} + (z-1) \sum_{k=2}^{K} \ket{u_{k}} \bra{u_{k}} \mat{D} \ket{1}. \label{RandomSERIdentity}
\mat{F} \mat{D} \vect{1} = \frac{z - 1 + K}{K} \vect{1} + (z-1) \sum_{k=2}^{K} \vect{u}_{k}\vect{u}_{k}^{\top} \mat{D} \vect{1}. \label{RandomSERIdentity}
\end{align}
Because the second term in Eq.~(\ref{RandomSERIdentity}) vanishes when the group sizes are equal, the exact probability distribution can be derived as follows: 
\begin{align}
&P\left[ (N-1) \mathcal{C} = m \right] \notag\\
%&= K^{-N} \oint \frac{dz}{2\pi i} z^{-(1+m)} \bra{1} \mat{F}^{N-1} \ket{1} \notag\\
&= \frac{1}{K^{N-1}} \oint \frac{dz}{2\pi i} z^{-(1+m)} (z-1+K)^{N-1} \notag\\
&= \frac{1}{K^{N-1}} \oint \frac{dz}{2\pi i} z^{-(1+m)} \sum_{k=0}^{N-1} \binom{N-1}{k} z^{k} (K-1)^{N-1-k} \notag\\
%&= \left(1-\frac{1}{K}\right)^{N-1} \binom{N-1}{m} \frac{1}{(K-1)^{m}} \notag\\
&= \binom{N-1}{m} \left(\frac{1}{K}\right)^{m} \left(1-\frac{1}{K}\right)^{N-1-m}. 
\label{RandomSERProb2}
\end{align}
Equivalently, 
\begin{align}
&P\left( \mathcal{C} \right) 
= \binom{N-1}{(N-1)\mathcal{C}} \left(\frac{1}{K}\right)^{(N-1)\mathcal{C}} \left(1-\frac{1}{K}\right)^{(N-1) (1-\mathcal{C})}.
\label{RandomSERProb3}
\end{align}
Therefore, $(N-1) \mathcal{C}$ follows a binomial distribution. 
This result can be interpreted as follows. 
We suppose that there are $N$ elements that are linearly aligned, and we assign group labels from one end. 
As we focus only on the consecutive property of the group labels, the label of the first element can be arbitrary. 
For the next $N-1$ elements, the probability that the label is consecutive to the previous one is $1/K$, whereas the complement probability is $1-1/K$ because the group label can be arbitrary as long as it is not identical to the previous one. 
We sum over all possible patterns that have consecutive labels $m$ times to obtain $P\left[ \mathcal{C} = \frac{m}{N-1} \right]$. 

Even when the group sizes are not equal, Eq.~(\ref{RandomSERProb2}) is close to the actual distribution as long as the second term in Eq.~(\ref{RandomSERIdentity}) is negligible. 
When $N \gg 1$ and the size of each group is of constant order, i.e., $N/K = O(1)$, Eq.~(\ref{RandomSERProb2}) is well approximated as a Poisson distribution. 
Furthermore, when $N/K \gg 1$, the distribution is nearly normal. 
The distribution of $\Delta\left( \vect{\pi}, \vect{\sigma} \right)$ is obtained by shifting the distribution (\ref{RandomSERProb3}) by a constant factor.

%%%%%%%%%%%%%%%%%%%%%%%%%%%%%%%%%%%%%%%%%%%%%%%%%%%%%%%%%%%
%%%%%%%%%%%%%%%%%%%%%%%%%%%%%%%%%%%%%%%%%%%%%%%%%%%%%%%%%%%

\section{Label continuity errors for nested partitions}\label{sec:LCENested}
\begin{figure}[t!]
  \centering
  \includegraphics[width= 0.99\columnwidth]{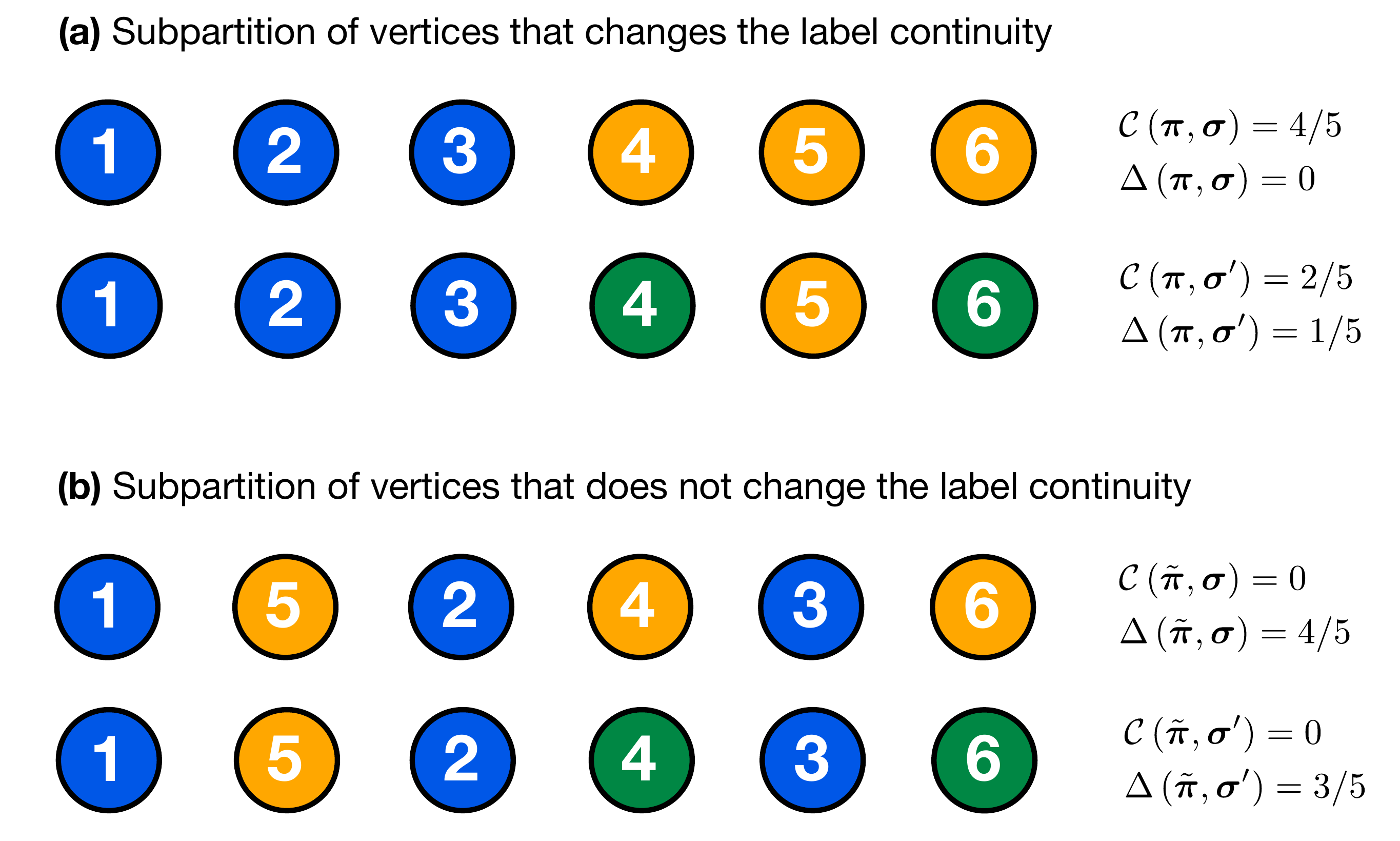}
  \caption{
  Effect of subpartitioning on the label continuity $\mathcal{C}$ and label continuity error $\Delta$. 
  The partition $\vect{\sigma}^{\prime}$ is a nested partition of $\vect{\sigma}$; the yellow label in $\vect{\sigma}$ is subpartitioned into the yellow and green labels in $\vect{\sigma}^{\prime}$. 
  (a) When the group labels are maximally consecutive (sequence $\vect{\pi}$), $\mathcal{C}$ becomes smaller by subpartitioning. 
  (b) When the group labels are not consecutive at all (sequence $\tilde{\vect{\pi}}$), $\mathcal{C}$ does not change regardless of the choice of the nested partition. 
  Although $\vect{\sigma}$ and $\vect{\sigma}^{\prime}$ are the same between (a) and (b), the value of $\mathcal{C}$ is affected by the locations of the blue labels.
	}
  \label{fig:SchematicLCEnested}
\end{figure}

As an example of partitions with different numbers of groups, here we investigate the difference in the {\LCE}s between a partition $\vect{\sigma}$ with $K$ groups and its nested partition $\vect{\sigma}^{\prime}$. 
The partition $\vect{\sigma}^{\prime}$ is obtained by subpartitioning the vertices $V_{K}$ having $K$th group label in $\vect{\sigma}$ into $V_{K,1}$ and $V_{K,2}$ ($V_{K,1} \cup V_{K,2} = V_{K}$); we denote the sizes of these two groups as $N^{\prime}_{K,1}$ and $N^{\prime}_{K,2}$ ($N^{\prime}_{K,1} + N^{\prime}_{K,2} = N_{K}$), and also denote $\{ N^{\prime}_{k} \} = \{ N_{1}, \dots, N_{K-1}, N^{\prime}_{K,1}, N^{\prime}_{K,2}\}$. 
The partitions $\vect{\sigma}$ and $\vect{\sigma}^{\prime}$ are only locally different. 
The difference in the {\LCE}s for $\vect{\sigma}$ and $\vect{\sigma}^{\prime}$ with the same sequence $\vect{\pi}$ is bounded as
\begin{align}
&- \frac{1}{N-1}
\le 
\Delta\left( \vect{\pi}, \vect{\sigma}^{\prime} \right) - \Delta\left( \vect{\pi}, \vect{\sigma} \right) \notag\\
&\le 
\frac{ 2 \min\{N^{\prime}_{K,1}, N^{\prime}_{K,2}\} - \delta\left( N^{\prime}_{K,1}, N^{\prime}_{K,2} \right) - 1}{N-1}. \label{NestedPartitionDifferenceBound}
\end{align}
The lower bound is trivial because the label continuity $\mathcal{C}$ is a nonnegative quantity and $\mathcal{C}$ cannot be smaller when $\mathcal{C}=0$ before subpartition. 
The upper bound of Eq.~(\ref{NestedPartitionDifferenceBound}) can be derived as follows.
The difference in the {\LCE} is maximized when the difference in $\mathcal{C}$ is maximized. 
Note that the number of flips of the labels can be maximized when the labels before the subpartition are aligned completely consecutively, e.g., the case in Fig.~\ref{fig:SchematicLCEnested}(a). 
In this case, we can maximize the difference in $\mathcal{C}$ by aligning the vertices in $V_{K,1}$ and $V_{K,2}$ as alternately as possible. 
The achieved difference is 
\begin{align}
\mathcal{C}\left( \vect{\pi}, \vect{\sigma}^{\prime} \right) - \mathcal{C}\left( \vect{\pi}, \vect{\sigma} \right) 
&= 
\begin{cases}
-\frac{N_{K}-1}{N-1} & (N^{\prime}_{K,1} = N^{\prime}_{K,2})\\
-\frac{ 2 \min\{N^{\prime}_{K,1}, N^{\prime}_{K,2}\} }{N-1} & (N^{\prime}_{K,1} \ne N^{\prime}_{K,2})
\end{cases} \notag\\
&= \frac{ \delta\left( N^{\prime}_{K,1}, N^{\prime}_{K,2} \right) - 2 \min\{N^{\prime}_{K,1}, N^{\prime}_{K,2}\} }{N-1}.
\end{align}

Equation (\ref{NestedPartitionDifferenceBound}) indicates that the {\LCE} is a local quantity, that is, the bound of variation in the {\LCE} is characterized by $N^{\prime}_{K,1}$ and $N^{\prime}_{K,2}$; the variation tends to be small when $\min\{N^{\prime}_{K,1}, N^{\prime}_{K,2}\}$ is small. 
However, when it comes to the specific difference, not bounds, it depends not only on the subsequence within $V_{K}$, but also on the position $V_{K}$ in the entire sequence $\vect{\pi}$ (see Fig.~\ref{fig:SchematicLCEnested} for specific examples). 
The present result implies that comparison of the {\LCE}s is generally complicated when partitions have different numbers of groups.

%%%%%%%%%%%%%%%%%%%%%%%%%%%%%%%%%%%%%%%%%%%%%%%%%%%%%%%%%%%
%%%%%%%%%%%%%%%%%%%%%%%%%%%%%%%%%%%%%%%%%%%%%%%%%%%%%%%%%%%

% If you have acknowledgments, this puts in the proper section head.
\begin{acknowledgments}
This work was supported by JST ACT-X Grant No. JPMJAX21A8 (Kawamoto), 
JSPS KAKENHI\ 19H01506, 22H00827 (Kawamoto and Kobayashi), 20H05633 (Kobayashi), and 
Quantum Science and Technology Fellowship Program (Q-STEP) (Ochi). 
\end{acknowledgments}

% Create the reference section using BibTeX:
%\bibliographystyle{unsrt}
%\bibliography{ref}

\seccountSI\label{sec:BetheHessianHyperparameter}%Sec S1

\tabcountSI\label{tab:dataset}%Tab S1

\figcountSI\label{fig:BetheHessianDependency}%Fig S1
\figcountSI\label{fig:MaxGroupSizeORGM}%Fig S2
\figcountSI\label{fig:phasediagramClusteringORGM}
\figcountSI\label{fig:empirical_adjacency_K2}
\figcountSI\label{fig:empirical_adjacency_K2_2}
\figcountSI\label{fig:empirical_adjacency_K2_3}
\figcountSI\label{fig:empirical_adjacency_K4}
\figcountSI\label{fig:karate_tau_K2}
\figcountSI\label{fig:polbooks_tau_K2}
\figcountSI\label{fig:norm_LCE_K2}
\figcountSI\label{fig:norm_LCE_K3}
\figcountSI\label{fig:norm_LCE_K4}
\figcountSI\label{fig:norm_LCE_K5}
\figcountSI\label{fig:norm_LCE_K6}

\clearpage

\setcounter{section}{0}
\setcounter{table}{0}
\setcounter{equation}{0}
\setcounter{figure}{0}
\setcounter{page}{1}
     
\renewcommand{\thetable}{S\arabic{table}}
\renewcommand{\thefigure}{S\arabic{figure}}
\renewcommand{\thesection}{S\arabic{section}}
\renewcommand{\theequation}{S\arabic{equation}}

\renewcommand\appendixname{} 
\begin{widetext}

{\flushleft
{\fontsize{16pt}{16pt}\selectfont
 \textbf{Supplementary Materials} \\
 \vspace{.7cm}
 \Large{``\papertitle''} \\
 \vspace{.5cm}
{\large Tatsuro Kawamoto, Masaki Ochi and Teruyoshi Kobayashi}
}
}

\vspace{2cm}

%%%%%%%%%%%%%%%%%%%%%%%%%%%%%%%%%%%%%%%%%%%%%%%%%%%%%%%%%%%
%%%%%%%%%%%%%%%%%%%%%%%%%%%%%%%%%%%%%%%%%%%%%%%%%%%%%%%%%%%

\section{Hyperparameter dependency in the Bethe Hessian}\label{sec:BetheHessianHyperparameter}
This section investigates the dependency of the hyperparameter $r$ and the choice of eigenvector in the Bethe Hessian $\mat{B}$ on spectral ordering.
Figure \ref{fig:BetheHessianDependency} shows, for each $\hat{\vect{\pi}}$ estimated by $k$th ($k=1,2,3,4$) eigenvector, the achieved value of $H_{2}\left( \hat{\vect{\pi}}; \mat{A} \right)$ as we sweep the hyperparameter $r$. 
In this experiment, we used an instance of the {\ORGM} with $N=100$, $c=6$, $\epsilon=0.1$, and $r/N=0.1$. 
The dashed line in the figure represents the default value of $r$. 

This result indicates that the eigenvector with which $H_{2}\left( \hat{\vect{\pi}}; \mat{A} \right)$ is minimized varies as $r$ increases, particularly when $r$ is relatively small. 
However, when $r$ is sufficiently large, the estimate $\hat{\vect{\pi}}$ using the eigenvector $\vect{\nu}_{2}$ associated with the second-smallest eigenvalue yields the lowest value of $H_{2}\left( \hat{\vect{\pi}}; \mat{A} \right)$. 

Based on this observation, we employ $\vect{\nu}_{2}$ for ordering with the Bethe Hessian. 
It should also be noted that, as shown in Fig.~\ref{fig:BetheHessianDependency}, the global minimum of $H_{2}\left( \hat{\vect{\pi}}; \mat{A} \right)$ is typically achieved when $r$ is lower than the default value we employed. 
Therefore, although it is not within the scope of this study, a better performance would be obtained if we optimize with respect to $r$.

%%%%%%%%%%%%%%%%%%%%%%%%%%%%%%%%%%%%%%%%%%%%%%%%%%%%%%%%%%%
%%%%%%%%%%%%%%%%%%%%%%%%%%%%%%%%%%%%%%%%%%%%%%%%%%%%%%%%%%%

\begin{figure}[tbh]
  \centering
  \includegraphics[width= 0.5\columnwidth]{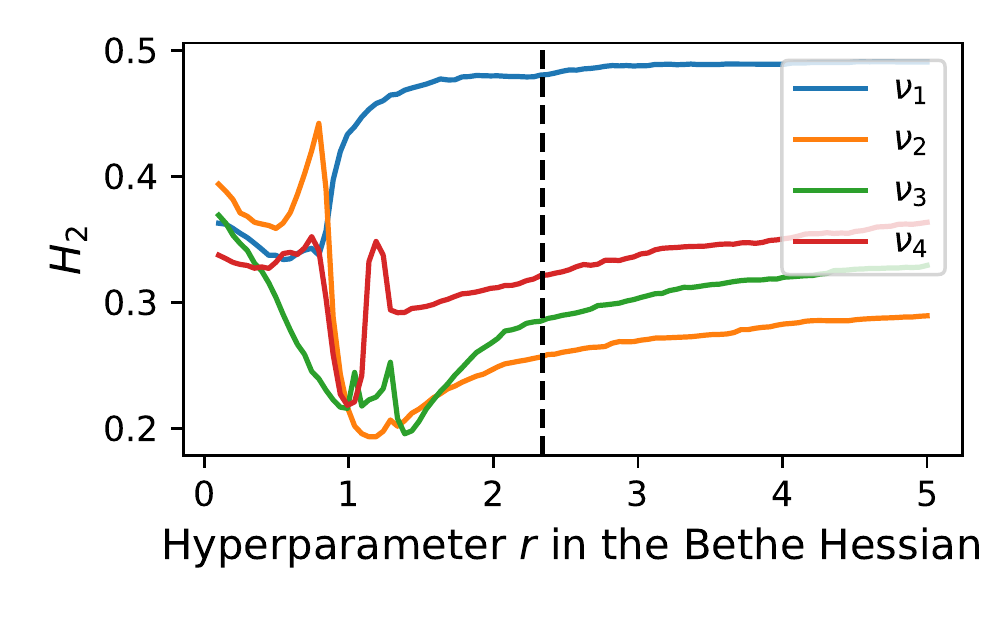}
  \caption{
  Dependency of the hyperparameter $r$ in the Bethe Hessian on the achieved value of $H_{2}\left( \hat{\vect{\pi}}; \mat{A} \right)$. 
  Each line represents the result based on the sequence $\hat{\vect{\pi}}$ estimated through the $k$th ($k=1,2,3,4$) eigenvector $\vect{\nu}_{k}$. 
	}
  \label{fig:BetheHessianDependency}
\end{figure}

\begin{table}[tbh]
\caption{Description of empirical datasets. All the data can be loaded from graph-tool~\cite{graph-tool}.}
\begin{center}
\begin{tabularx}{\textwidth}{lcclc}
 \hline
   \multirow{2}{*}{Dataset} &   \multirow{2}{*}{$N$} &   \multirow{2}{*}{$M$}  &\multirow{2}{*}{Data description} & \multirow{2}{*}{Reference} \\
   & & & &  \\
   \hline 
           adjnoun  &  112 &   425  & Word adjacencies of common adjectives and nouns in the novel \emph{David Copperfield}. &  \cite{newman2006finding} \\             
           celegans &  297 &  2,359  & Neural connections of the C.\ elegans nematode. &       \cite{white1986structure} \\  
          dolphins &  62    & 159    &  Frequent associations among dolphins  in a community.  &  \cite{lusseau2003dolphins}  \\
           football    &  115 &   613 &  Network of American football games between Division IA colleges. &   \cite{girvan2002community}, \cite{evans2010clique} \\  
          karate    &  34 &   77 &  Network of friendships among members of a university karate club.  &   \cite{zachary1977karate} \\  
          lesmis    &  77 &   254 &   Character co-appearance network of  \emph{Les Mis\'{e}rables}. & \cite{knuth1993stanford}  \\  
          netscience    &  1,589 &  2,742 &  Collaboration network among scientists working on network science.  &   \cite{newman2006finding} \\  
          polblogs &  1,490 &   19,090  &  Network of hyperlinks among U.S. political blogs. &       \cite{adamic2005polblogs}   \\                       
          polbooks &  105 &   441  &  Co-purchase network of books on US politics. &       \cite{polbook}   \\                       
          foodweb &  161 &   592  &  Plant and mammal food web in the Serengeti savanna ecosystem.  &       \cite{baskerville2011serengeti}   \\               
           \hline 
 \end{tabularx}
 \end{center}
 \label{tab:dataset}
\end{table}

\begin{figure*}[t!]
  \centering
  \includegraphics[width= \columnwidth]{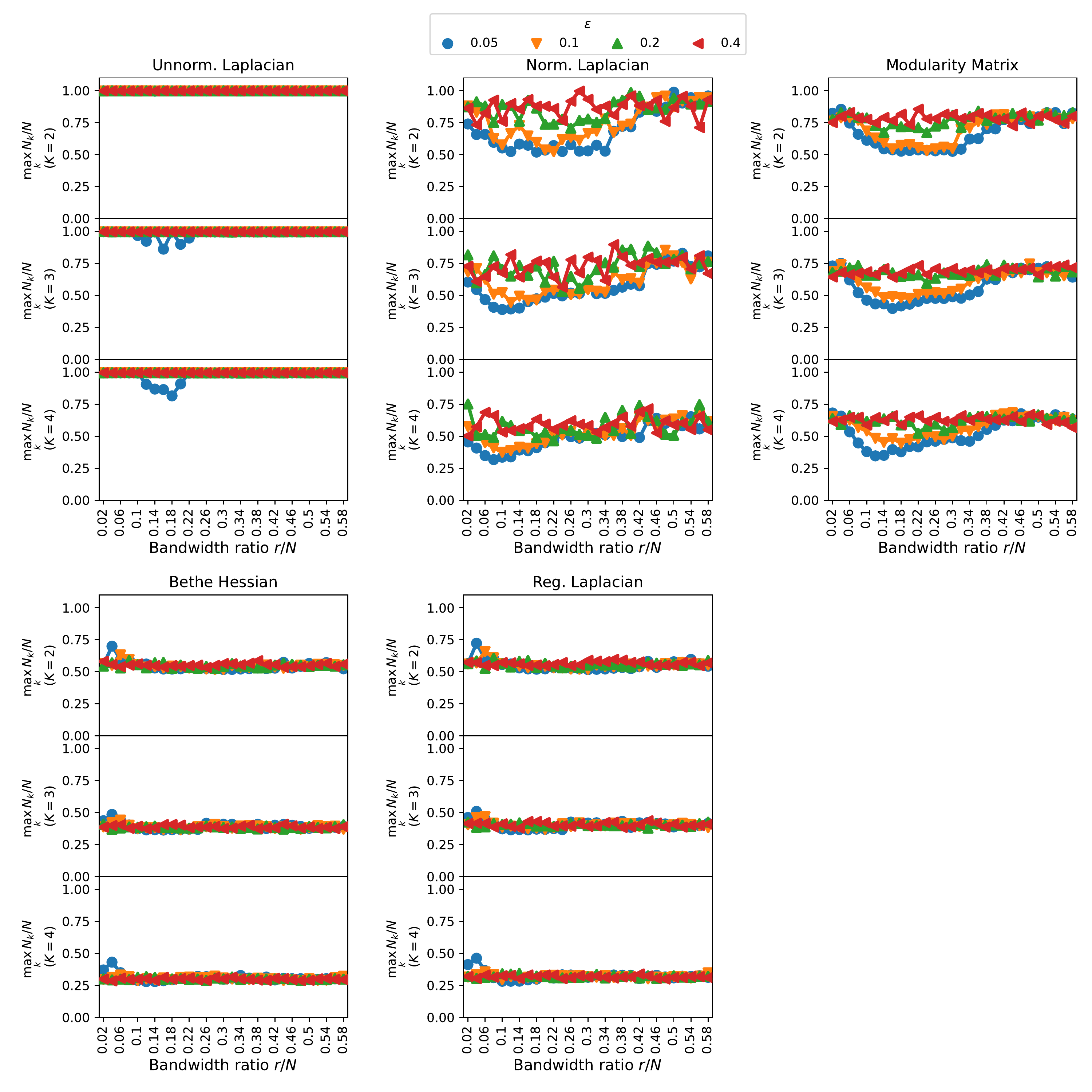}
  \caption{
  Performance of the spectral clustering method on graphs generated by the {\ORGM}. 
  The graphs are generated by the {\ORGM} with $N=1,000$ and $c=6$. 
  Each panel shows the fraction of the largest group $\max_{k} \{N_{k}\}/N$ for various parameter sets of the {\ORGM} in each matrix used in the spectral clustering. 
  Each point represents the $10$-sample average of $\max_{k} \{N_{k}\}/N$ under the same parameter set. 
	}
  \label{fig:MaxGroupSizeORGM}
\end{figure*}

\begin{figure*}[t!]
  \centering
  \includegraphics[width= 0.82\columnwidth]{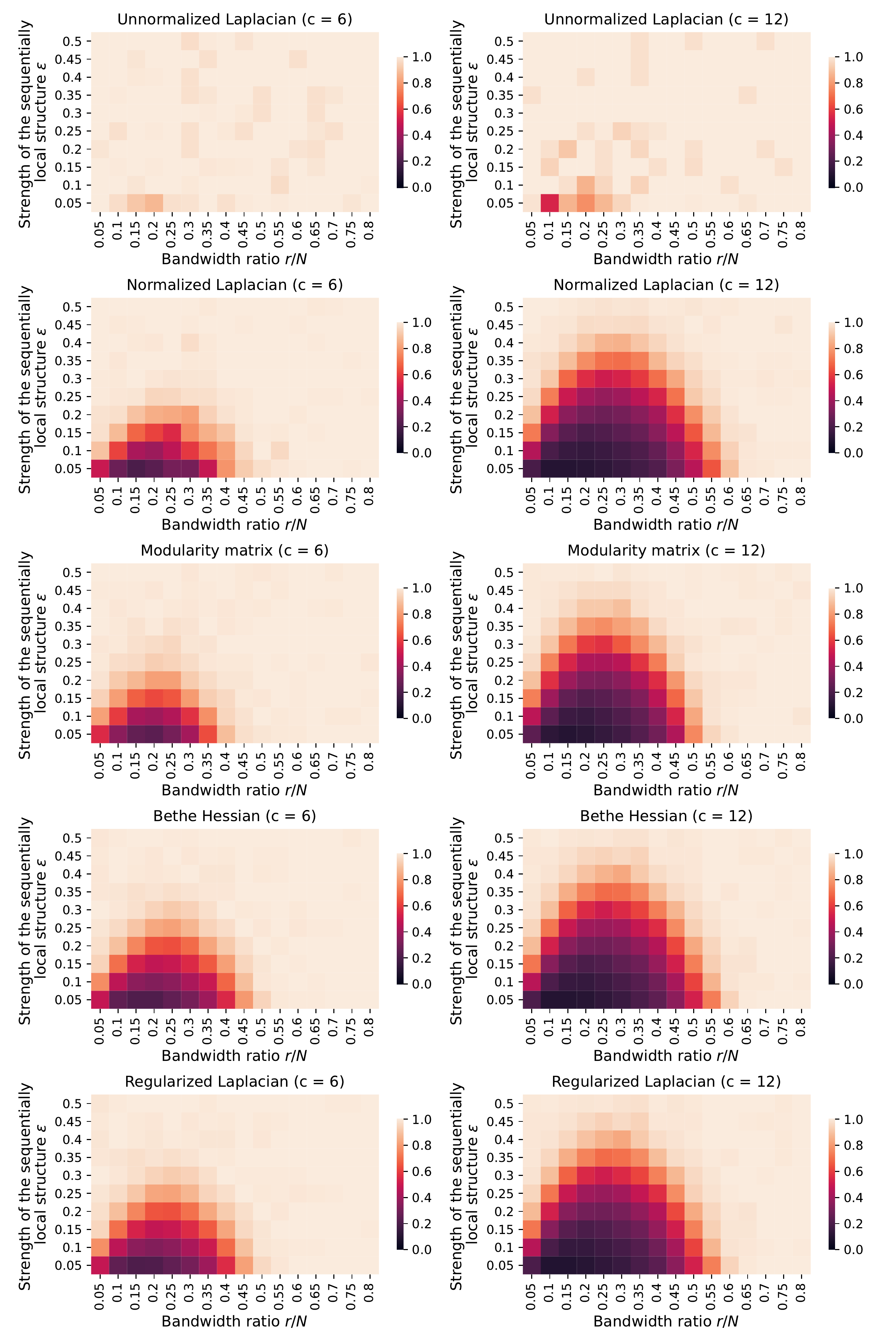}
  \caption{
  Phase diagrams of the normalized {\LCE}s in the $(r/N, \epsilon)$-space. 
  For graphs generated by the {\ORGM} with $N=1,000$, we conducted the spectral clustering methods with $K=2$ and measured the normalized {\LCE}. 
  Each point represents the $30$-sample average of the normalized {\LCE} under the same parameter set. 
  We set the average degrees to (a) $c=6$ and (b) $c=12$. 
	}
  \label{fig:phasediagramClusteringORGM}
\end{figure*}

\begin{figure}[thb]
  \centering
  \includegraphics[width= 15cm]{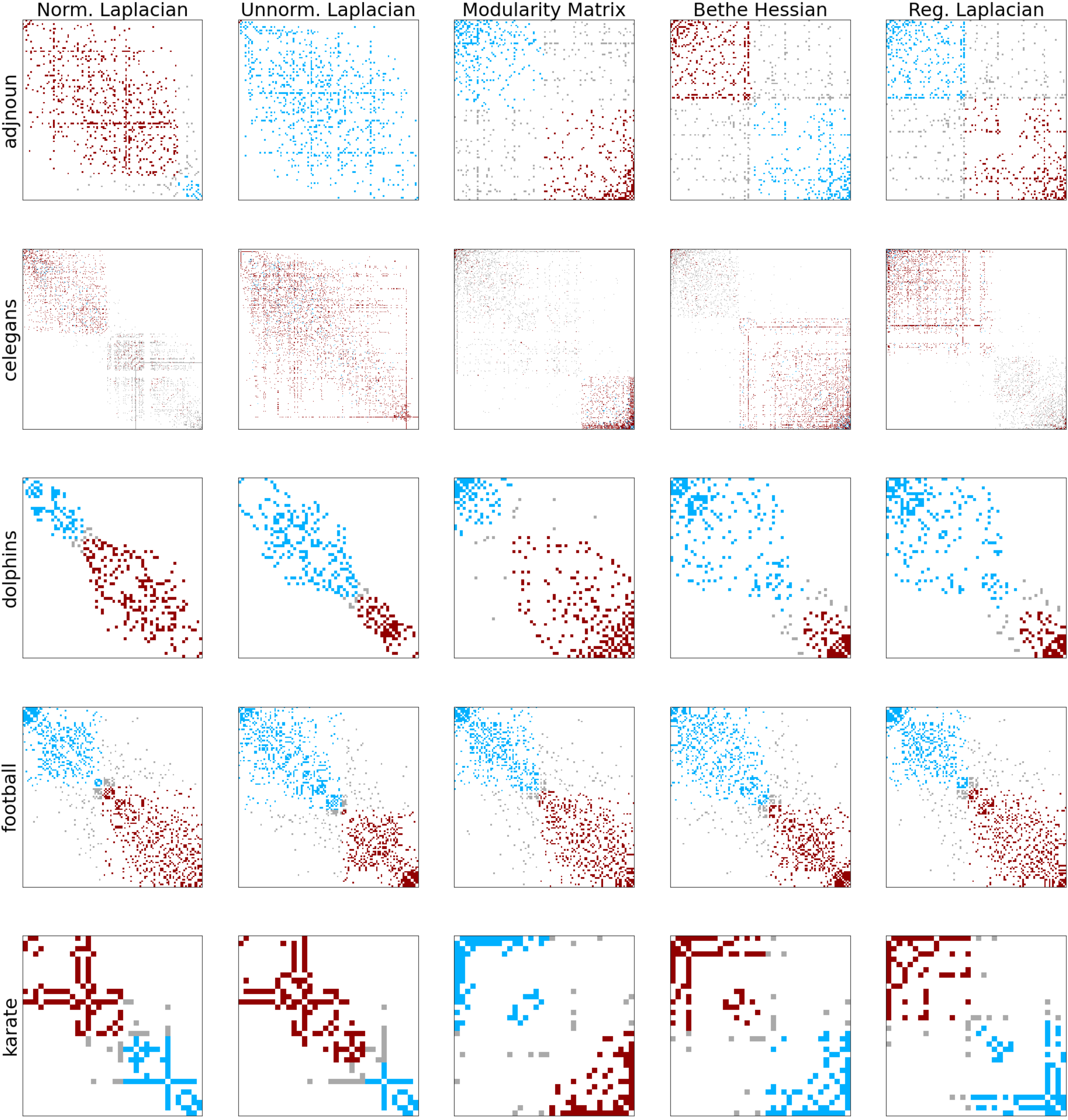}
  \caption{Adjacency matrix with vertices being aligned with the spectral ordering methods: $K=2$. 
         Colors denote vertex groups inferred by the K-means method. 
         Edges between vertices in different groups are colored in gray.
	}
  \label{fig:empirical_adjacency_K2}
\end{figure}

\begin{figure}[thb]
  \centering
  \includegraphics[width= 15cm]{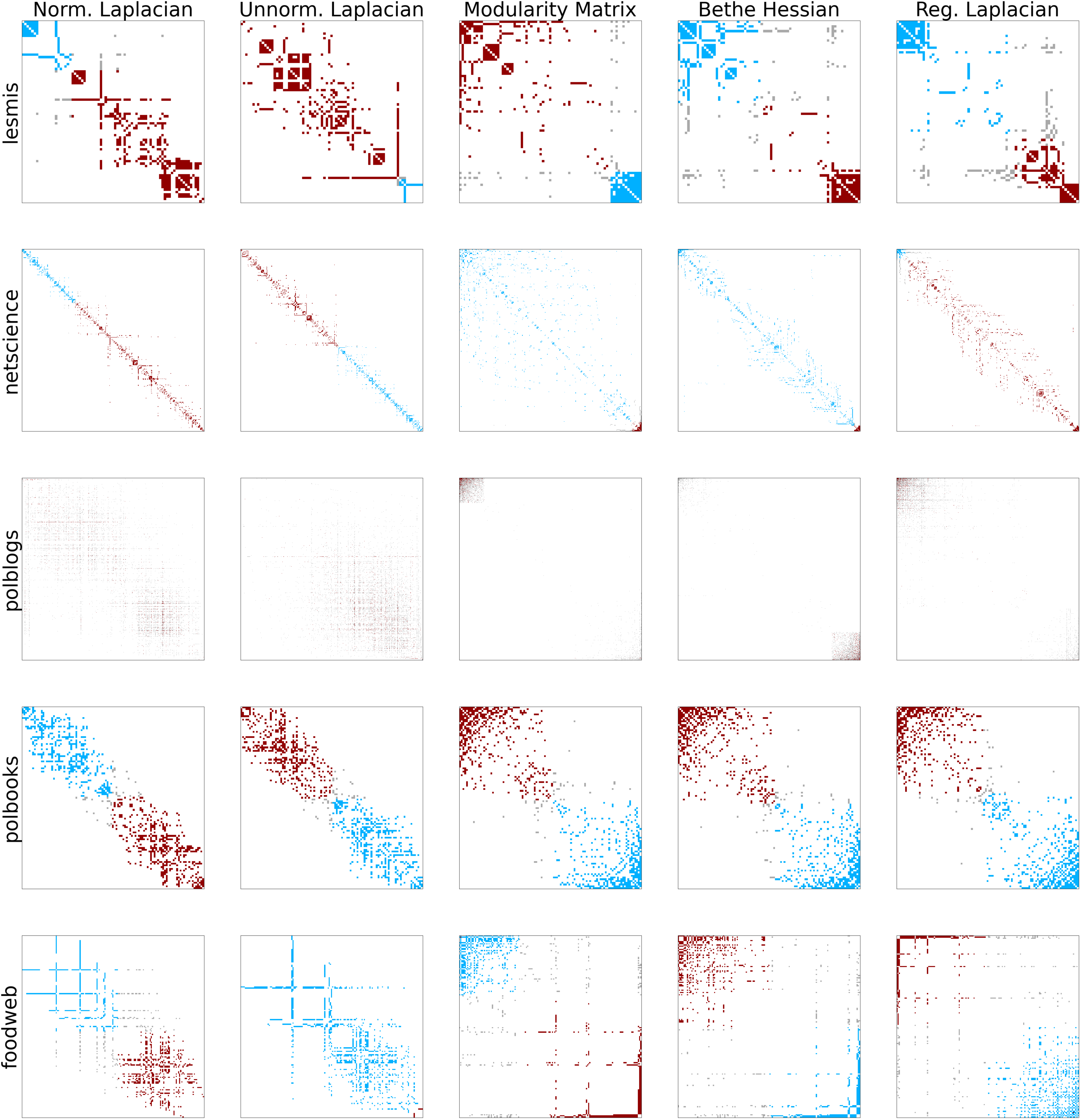}
  \caption{Adjacency matrix with vertices being aligned with the spectral ordering methods: $K=2$. 
         See the caption of Fig.~\ref{fig:empirical_adjacency_K2} for details.
	}
\end{figure}

\begin{figure}[thb]
  \centering
  \includegraphics[width= 15cm]{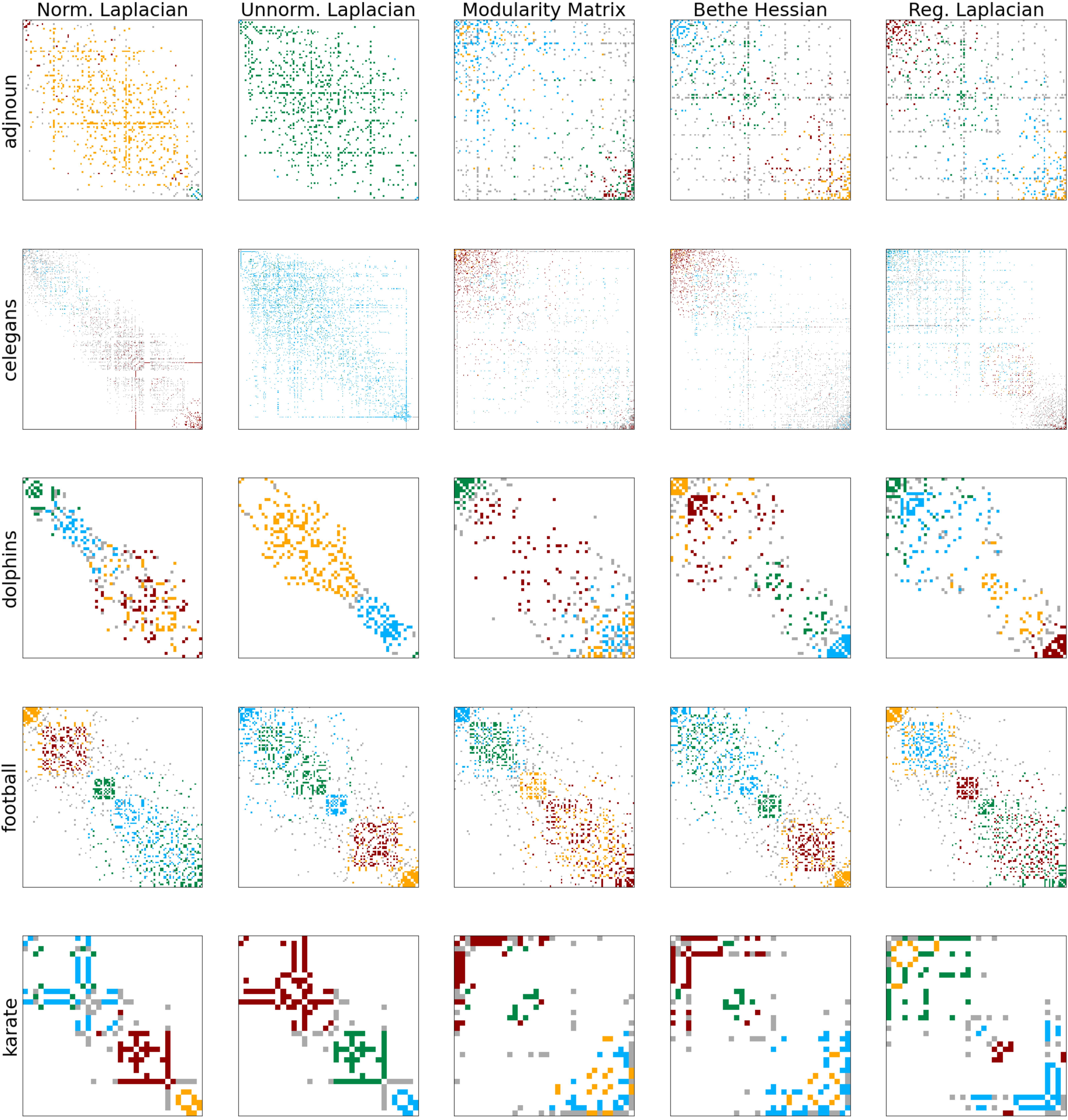}
  \caption{Adjacency matrix with vertices being aligned with the spectral ordering methods: $K=4$. 
         See the caption of Fig.~\ref{fig:empirical_adjacency_K2} for details.
	}
\end{figure}

\begin{figure}[thb]
  \centering
  \includegraphics[width= 15cm]{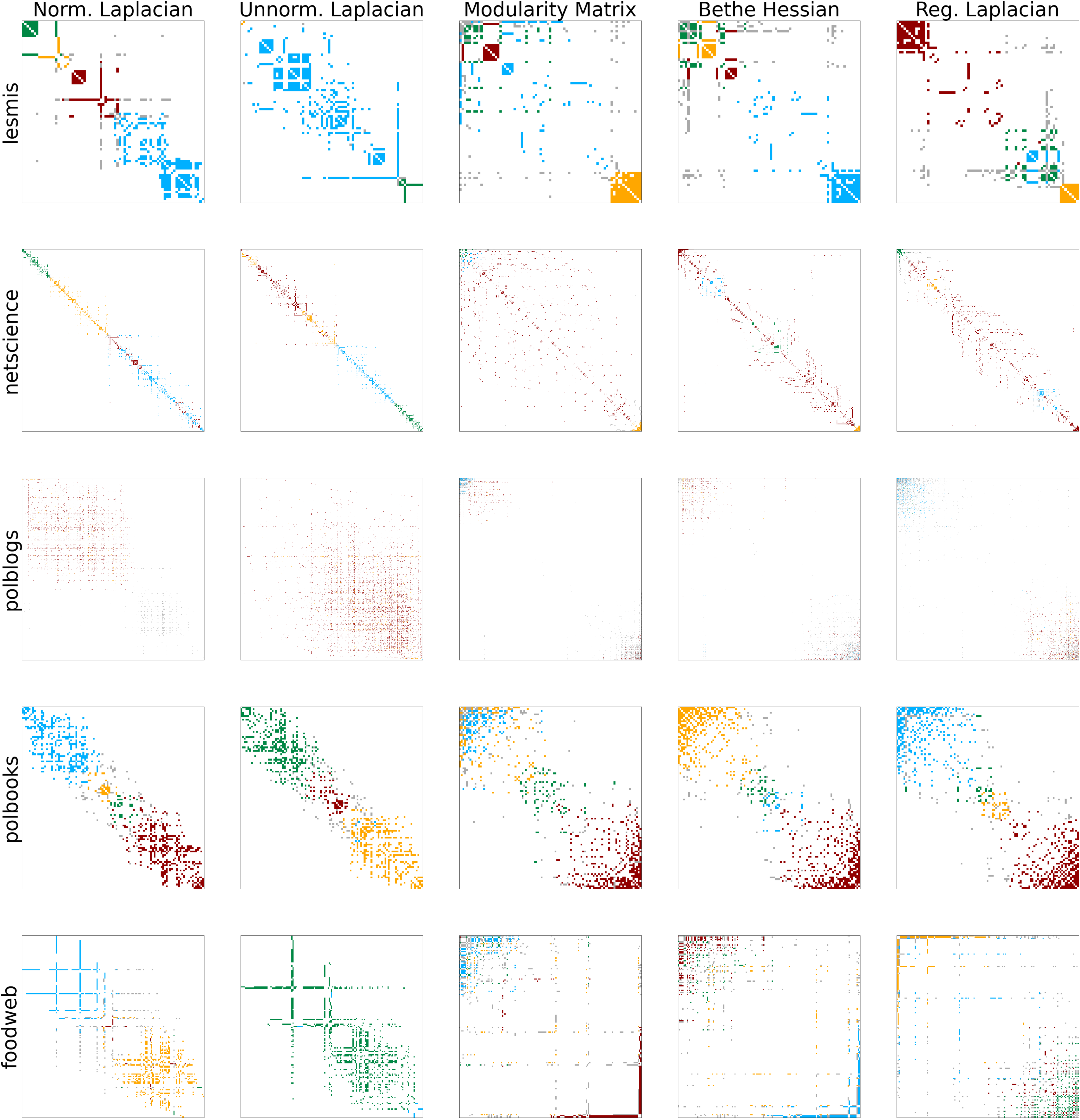}
  \caption{Adjacency matrix with vertices being aligned with the spectral ordering methods: $K=4$. 
         See the caption of Fig.~\ref{fig:empirical_adjacency_K2} for details.
	}
  \label{fig:empirical_adjacency_K4}
\end{figure}

\clearpage

\begin{figure}[thb]
  \centering
  \includegraphics[width= 11cm, angle=-90]{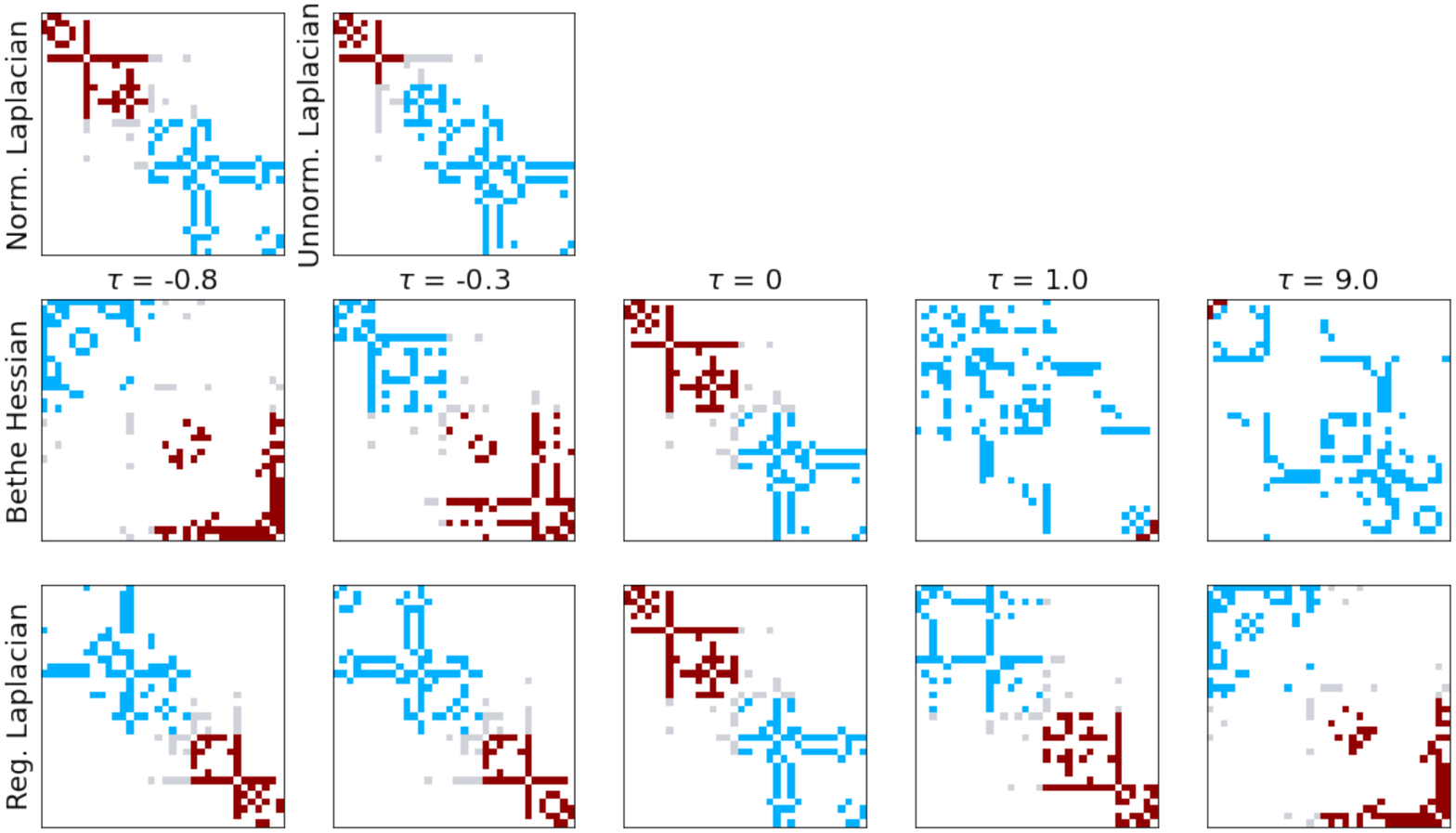}
  \caption{Effect of regularization parameter $\tau$ on the spectral ordering based on Bethe Hessian and regularized Laplacian: the Karate club data ($K=2$).  
        $\tau = -0.8,-0.3,0,1$, and $9$ respectively corresponds to $r = 5, 1.5,1,0.5$, and $0.1$.
        Note that for $\tau=0$ and $r=1$, the optimized sequences for Bethe Hessian and regularized Laplacian are almost identical to that of normalized Laplacian.
         Colors denote vertex groups inferred by the K-means method. 
         Edges between vertices in different groups are colored in gray.
	}
  \label{fig:karate_tau_K2}
\end{figure}

\begin{figure}[thb]
  \centering
  \includegraphics[width= 11cm, angle=-90]{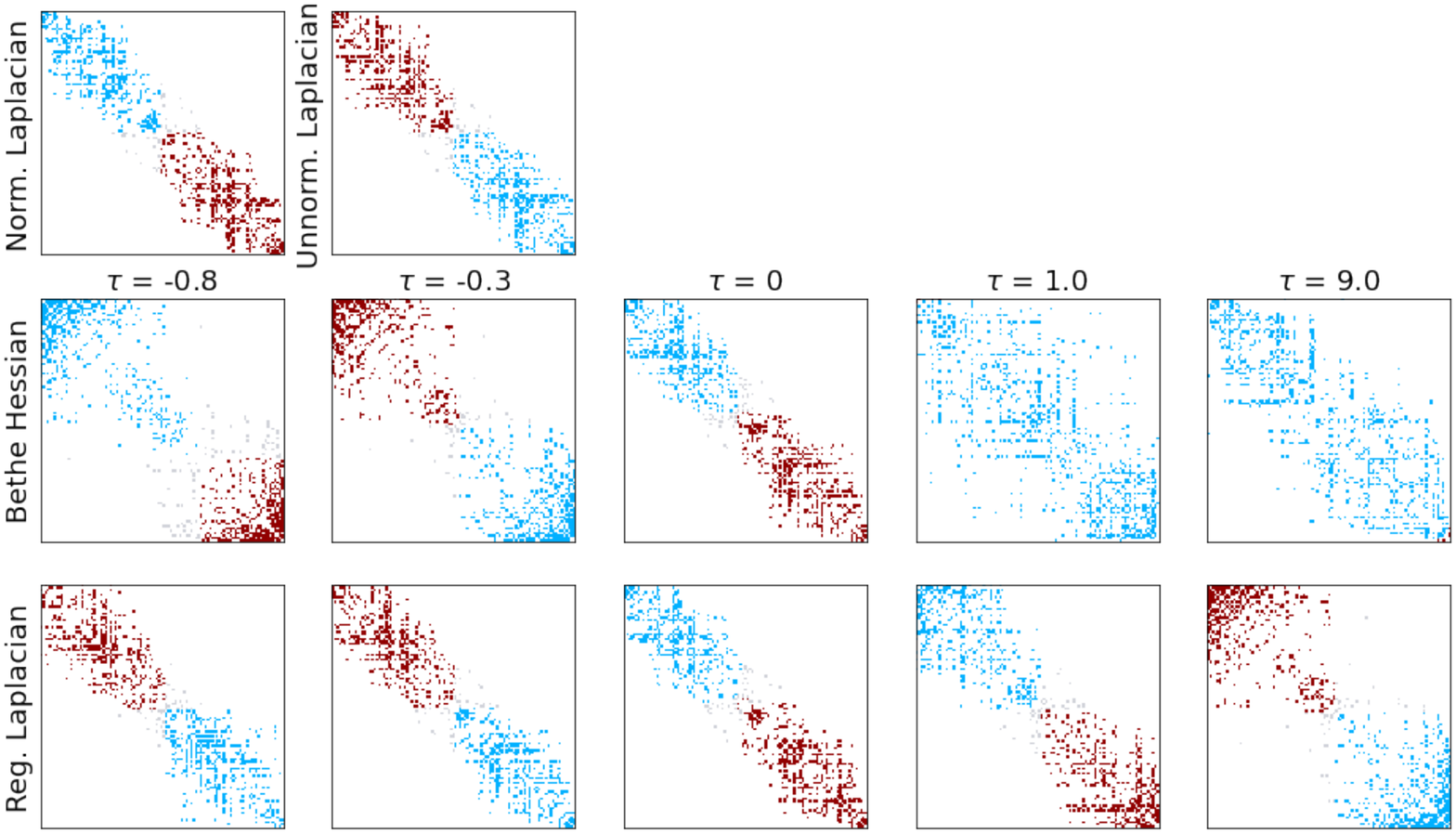}
  \caption{Effect of regularization parameter $\tau$ on the spectral ordering based on Bethe Hessian and regularized Laplacian: the political books data ($K=2$).  
     See the caption of Fig.~\ref{fig:karate_tau_K2} for details. 
	}
  \label{fig:polbooks_tau_K2}
\end{figure}

\clearpage

\begin{figure}[tbp]
  \begin{minipage}[b]{0.48\columnwidth}
    \centering
  \includegraphics[width=\columnwidth]{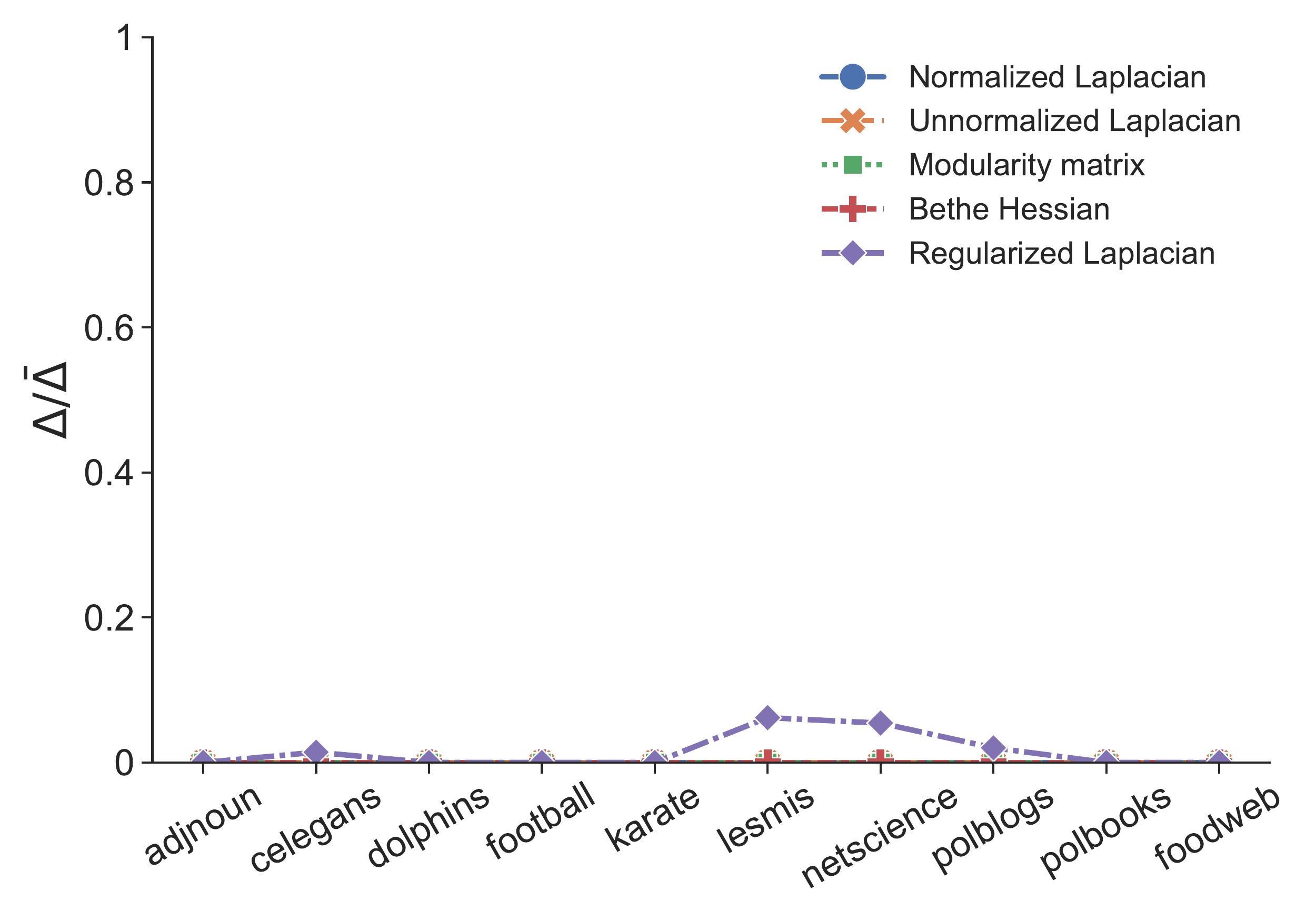}
  \caption{Normalized LCE for $K=2$.
	}
  \label{fig:norm_LCE_K2}
  \end{minipage}
  \begin{minipage}[b]{0.48\columnwidth}
    \centering
  \includegraphics[width=\columnwidth]{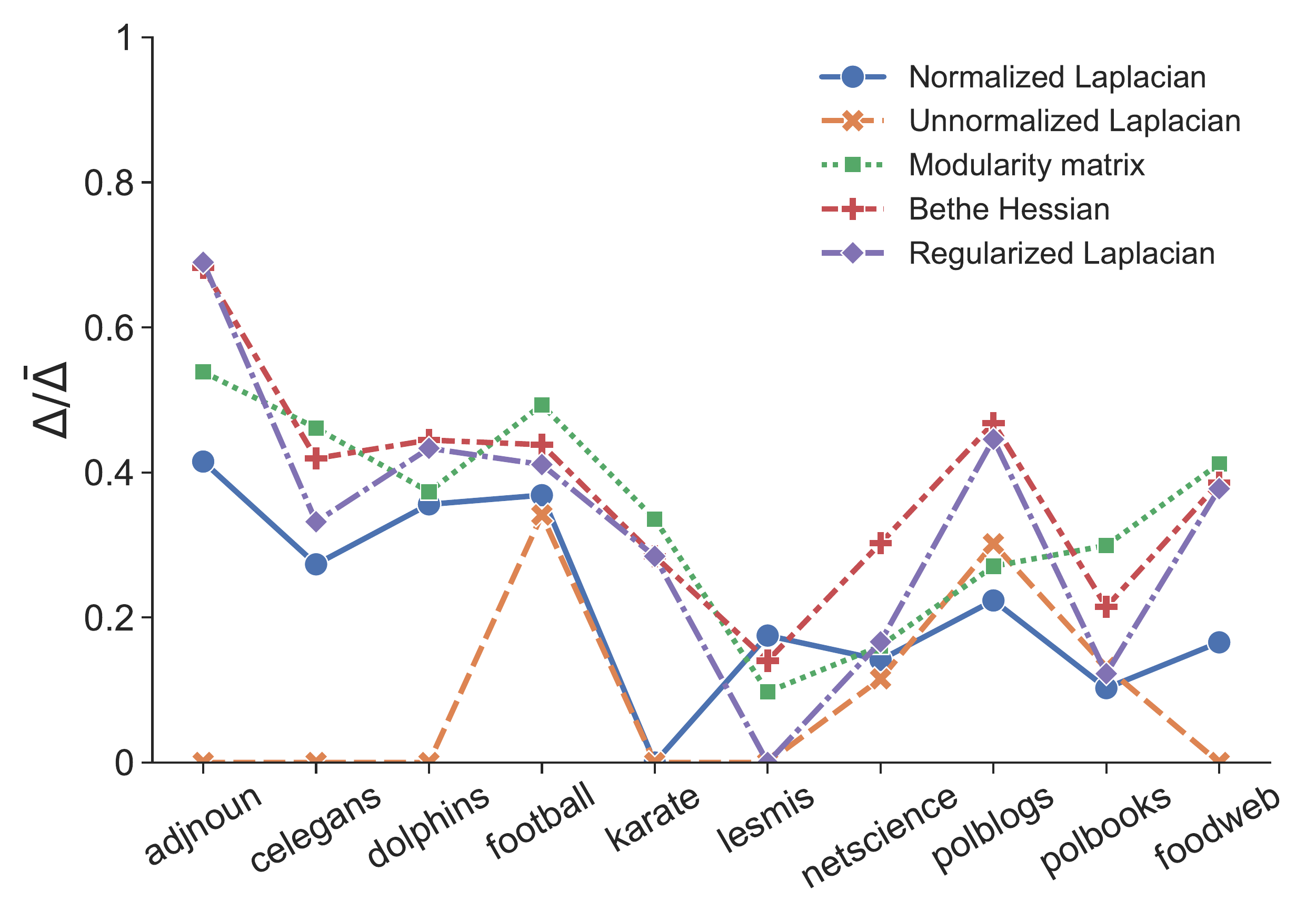}
  \caption{Normalized LCE for $K=3$.
	}
  \label{fig:norm_LCE_K3}
  \end{minipage}
%  \hspace{0.04\columnwidth} % \UTF{00E3}\UTF{0081}\UTF{0093}\UTF{00E3}\UTF{0081}\UTF{0093}\UTF{00E3}\UTF{0081}$B!x(B\UTF{00E9}\UTF{009A}\UTF{0099}\UTF{00E9}\UTF{0096}\UTF{0093}\UTF{00E4}\UTF{00BD}\UTF{009C}\UTF{00E6}\UTF{0088}\UTF{0090}%
  \begin{minipage}[b]{0.48\columnwidth}
    \centering
  \includegraphics[width=\columnwidth]{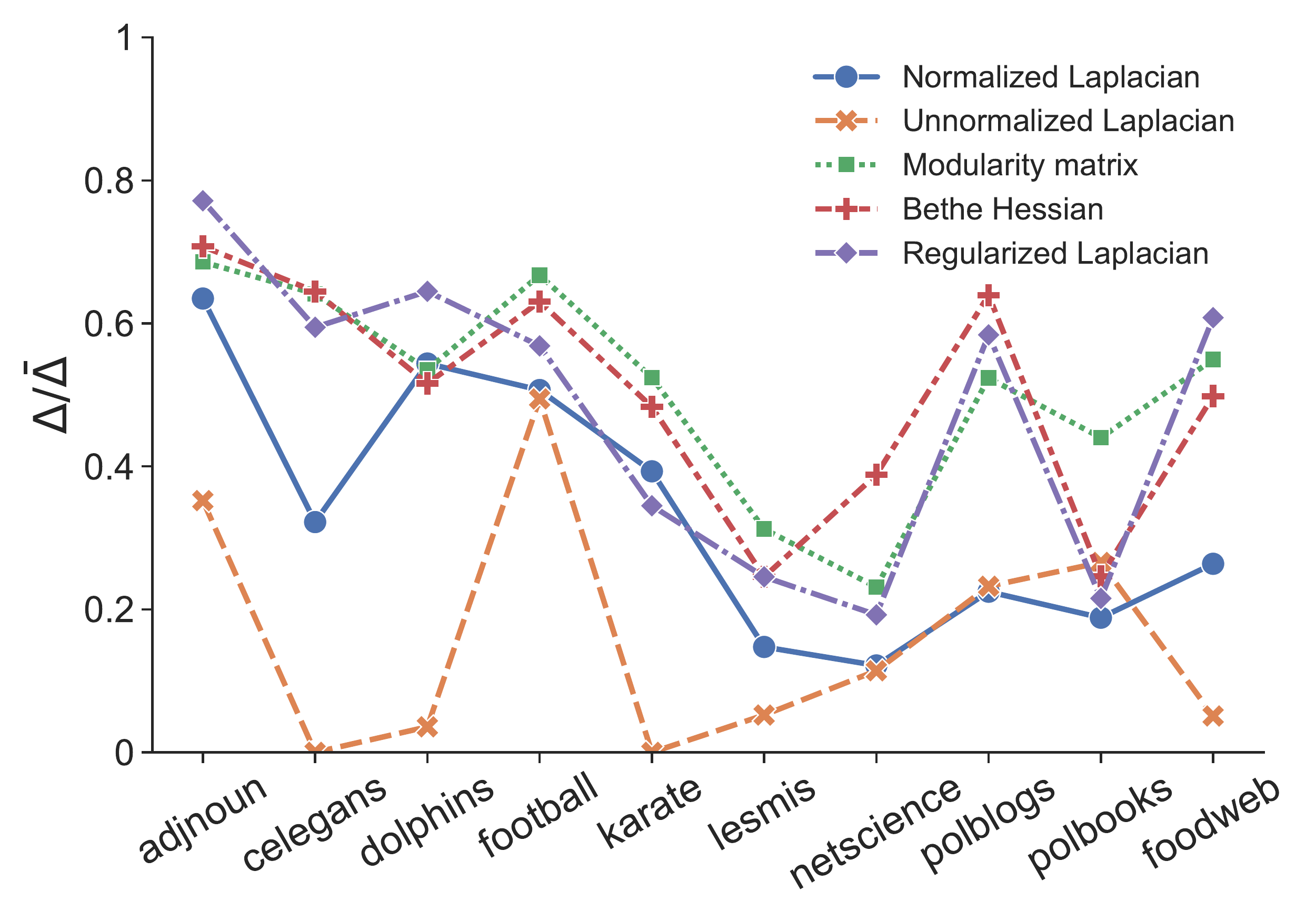}
  \caption{Normalized LCE for $K=4$.
	}
  \label{fig:norm_LCE_K4}
  \end{minipage}
%\end{figure}

%\begin{figure}[tbp]
  \begin{minipage}[b]{0.48\columnwidth}
    \centering
  \includegraphics[width=\columnwidth]{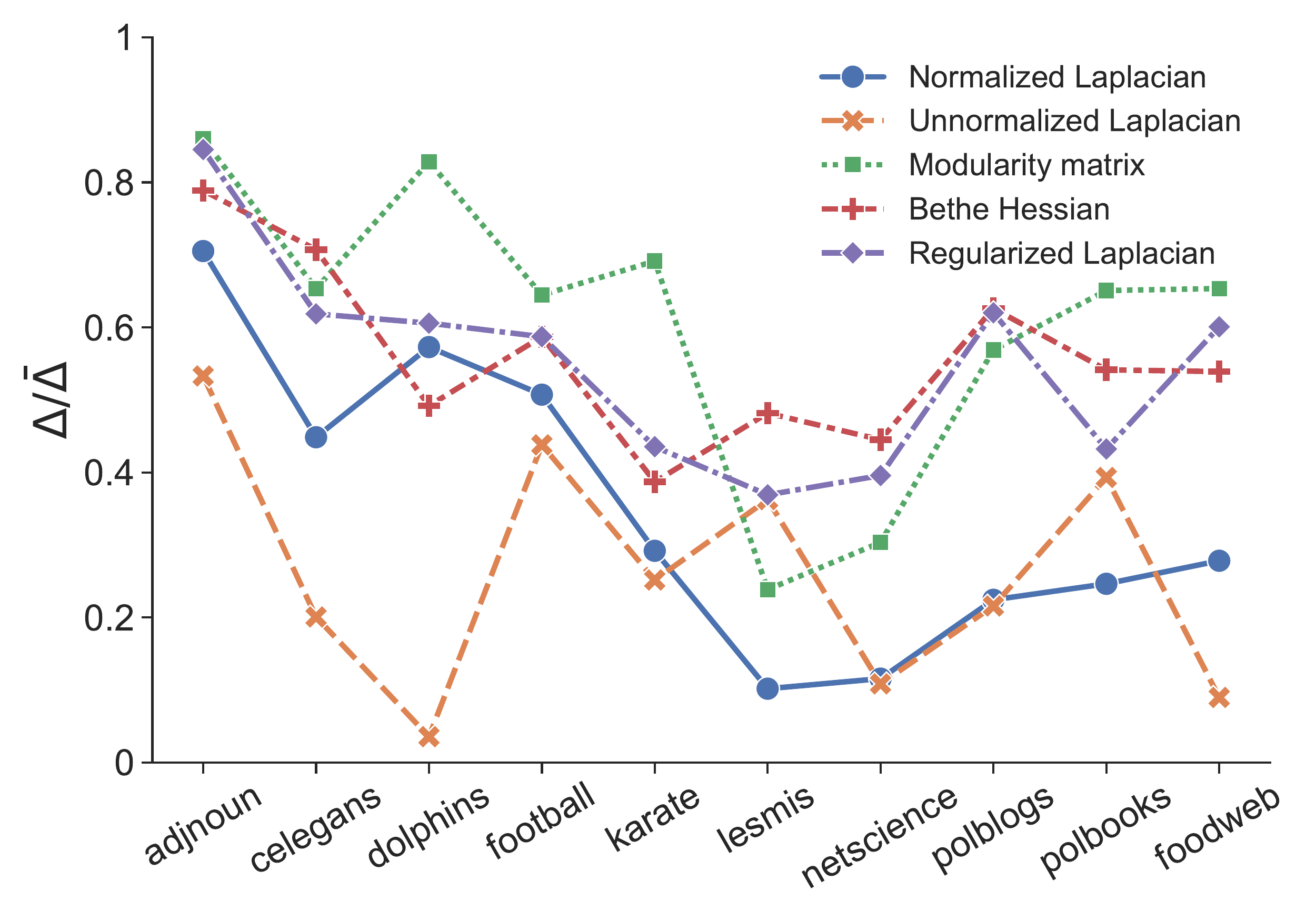}
  \caption{Normalized LCE for $K=5$.
	}
  \label{fig:norm_LCE_K5}
  \end{minipage}
%  \hspace{0.04\columnwidth} % \UTF{00E3}\UTF{0081}\UTF{0093}\UTF{00E3}\UTF{0081}\UTF{0093}\UTF{00E3}\UTF{0081}$B!x(B\UTF{00E9}\UTF{009A}\UTF{0099}\UTF{00E9}\UTF{0096}\UTF{0093}\UTF{00E4}\UTF{00BD}\UTF{009C}\UTF{00E6}\UTF{0088}\UTF{0090}%
  \begin{minipage}[b]{0.48\columnwidth}
    \centering
  \includegraphics[width=\columnwidth]{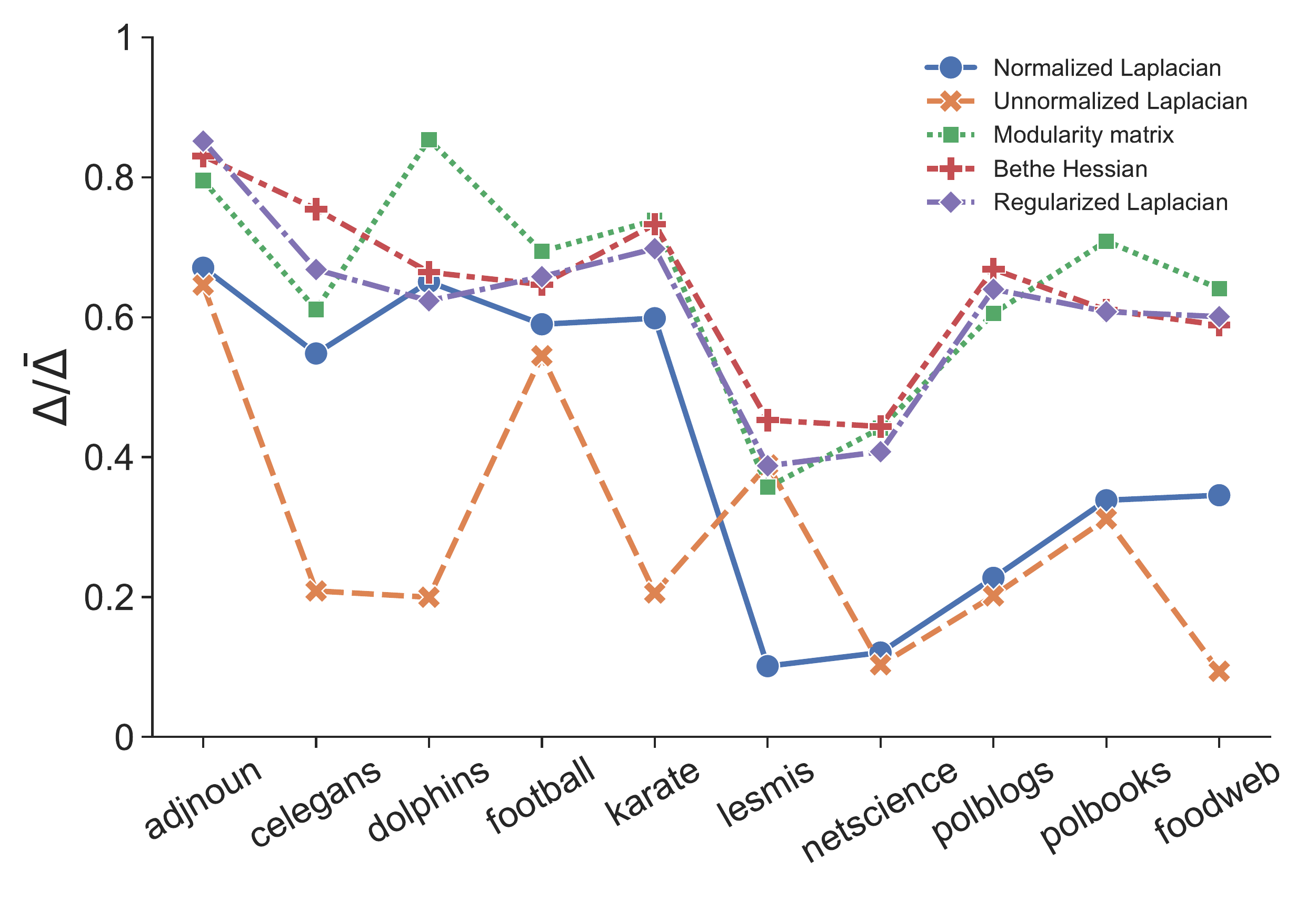}
  \caption{Normalized LCE for $K=6$.
	}
  \label{fig:norm_LCE_K6}
  \end{minipage}
\end{figure}

\end{widetext}

\end{document}